\documentclass[letterpaper, 10 pt, conference]{ieeeconf}  

\IEEEoverridecommandlockouts                              

\usepackage{amsfonts}		
\usepackage{amsmath}

\usepackage{bbm}  
\usepackage{dsfont}

\usepackage[pdftex]{graphicx}
\usepackage{booktabs} 
\usepackage{svg}
\usepackage{caption}
\usepackage{subcaption}
\usepackage{hyperref}

\usepackage{algpseudocode}
\usepackage{algorithm}
\usepackage{tabularx}

\makeatletter
\newcommand{\multiline}[1]{%
  \begin{tabularx}{\dimexpr\linewidth-\ALG@thistlm}[t]{@{}X@{}}
    #1
  \end{tabularx}
}
\makeatother

\newcommand{\PAR}[1]{\vskip2pt \noindent{\bf #1}}

\title{\LARGE \bf
A Multiplicative Value Function for \\ Safe and Efficient Reinforcement Learning
}

\author{Nick Bührer$^{1}$, Zhejun Zhang$^{1}$, Alexander Liniger$^{1}$, Fisher Yu$^{1}$, Luc Van Gool$^{1,2}$ 
\thanks{$^{1}$All authors are with the Computer Vision Lab, ETH Zurich, Switzerland. {\tt\footnotesize buehrern@ethz.ch,\,\{zhejun.zhang,alex.liniger,\newline vangool\}@vision.ee.ethz.ch,\,i@yf.io}}%
\thanks{$^{2}$Luc Van Gool is with PSI, KU Leuven, Belgium.}%
}

\begin{document}

\maketitle
\thispagestyle{empty}
\pagestyle{empty}

\begin{abstract}

An emerging field of sequential decision problems is safe Reinforcement Learning (RL), where the objective is to maximize the reward while obeying safety constraints. Being able to handle constraints is essential for deploying RL agents in real-world environments, where constraint violations can harm the agent and the environment. To this end, we propose a safe model-free RL algorithm with a novel multiplicative value function consisting of a safety critic and a reward critic. The safety critic predicts the probability of constraint violation and discounts the reward critic that only estimates constraint-free returns. By splitting responsibilities, we facilitate the learning task leading to increased sample efficiency. We integrate our approach into two popular RL algorithms, Proximal Policy Optimization and Soft Actor-Critic, and evaluate our method in four safety-focused environments, including classical RL benchmarks augmented with safety constraints and robot navigation tasks with images and raw Lidar scans as observations. Finally, we make the zero-shot sim-to-real transfer where a differential drive robot has to navigate through a cluttered room. Our code can be found at \url{https://github.com/nikeke19/Safe-Mult-RL}.

\end{abstract}


\section{Introduction}
\label{sec:intro}
Reinforcement Learning (RL) has made significant progress in recent years. Breakthroughs have been achieved on playing Atari~\cite{atari} and board games like Go~\cite{go}, as well as multi-agent RL on Starcraft~\cite{starcraft} and Dota~\cite{dota}. 
While some works like Miki et al.~\cite{anymal} deploy RL agents on real-world systems, most of the research is still conducted in simulation~\cite{gran_turismo,carla}. On the one hand, the sim-to-real transfer requires high-fidelity simulators and robust models. On the other hand, there is the safety aspect. In simulation, the agent can perform any action without real consequences. However, when robots are deployed in reality, not every action is admissible. This can be due to damage to the agent, e.g., when a robot crashes into an obstacle, or damage to the environment, e.g., when the obstacle is a human. Thus, it is necessary to consider safety by design. We can formulate safety requirements using Constrained Markov Decision Process (CMDP)~\cite{cmdp}, where in addition to maximizing the reward, the RL agent has to fulfill constraints on the expected accumulated safety cost. 

\begin{figure*}
    \centerline{\includegraphics[width=\textwidth]{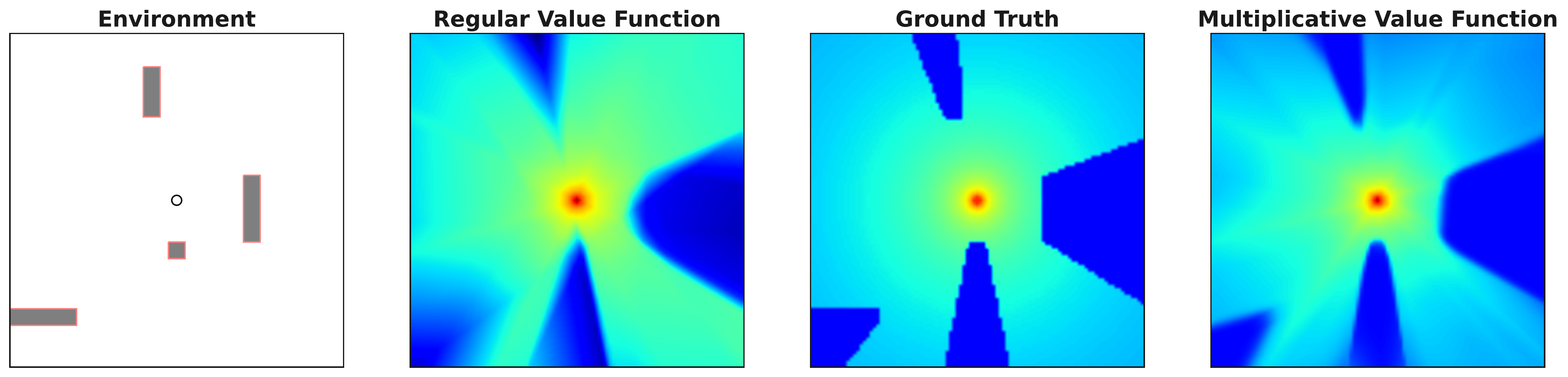}}
    \caption{Linear Quadratic Regulator policy evaluation of a point robot navigating towards the goal in the middle while avoiding the gray obstacles. The ground truth return shows sharp discontinuities that are better replicated by the multiplicative value function.}
    \label{fig:f_motivation}
    \vspace{-2ex}
\end{figure*}
    
Usually, the expected safety cost is estimated with a value function, called the safety critic. With a particular choice of safety cost, the cumulative cost constraint can be transformed into a chance constraint and be relaxed with a Lagrange multiplier. In contrast to previous works \cite{lagrange_old,lagrange_new, safety_transfer_learning,lagrange_pid}, we additionally propose a novel multiplicative value function where the safety critic explicitly addresses constraints and discounts a reward critic that only estimates constraint-free returns. 
The multiplicative value function has several advantages. A standard RL algorithm can learn safe behavior by specifying a penalty for constraint-violating actions. However, the performance can be sensitive to the magnitude of the penalty~\cite{cpo}. 
In contrast, we do not need to specify the magnitude: the reward critic neglects constraint violating returns, and the safety critic learns a binary decision. Moreover, penalties are often large in magnitude to discourage standard RL agents from constraint violating actions. As a result, there can be sharp discontinuities in the value landscape as shown in Fig.~\ref{fig:f_motivation}, which makes it difficult for a regular neural network to learn the value function. In our approach, the reward critic does not have to learn these discontinuities. Instead, the responsibility is shifted to the safety critic that estimates the probability of constraint violation. This makes the model optimization better behaved and leads to faster convergence with increased stability. 
    
We combine our approach with two popular algorithms: Proximal Policy Optimization (PPO)~\cite{ppo} and Soft Actor-Critic (SAC)~\cite{sac}, but in general, it can be combined with any on-policy or off-policy method that relies on a value or advantage function. To evaluate the effectiveness of our approach, we construct four safety-critical environments ranging from low to high dimensional observations based on images and Lidar scans. Finally, we deploy our algorithm on a real robot and perform map-less navigation. 
    


Our contributions are summarized as follows:

\begin{itemize}
      \item We introduce a novel multiplicative value function, that combines a regular value function and a safety critic in a multiplicative fashion. 
      We integrate this multiplicative value function into PPO and SAC.
      \item We test our methods on a suit of safety-critical environments and show that our methods outperform competing safe RL methods.
      \item We conduct experiments on a real-world robot navigation task in a cluttered environment and show zero-shot sim-to-real transfer. 
    \end{itemize}

\section{Related Work}




An overview of the recent advances in safe RL can be found in the latest surveys~\cite{liu2021policy,gu2022review}. 
In this section, we will focus on the prior works most related to our approach.

\textbf{Primal-dual approaches}, i.e., Lagrangian-based approaches, convert the CMDP into an unconstrained problem by relaxing the safety constraint with a Lagrange multiplier. 
This is the most straightforward way to solve a CMDP and is also the most related to our work. 
A Lagrange multiplier with heuristic update rules was first proposed in~\cite{lagrange_old}. More recent works established theoretical groundwork by proving convergence guarantees~\cite{lagrange_new, tessler2018reward} and zero duality gap~\cite{paternain2019constrained}. Practically, the Lagrangian approach has been integrated with PPO/TRPO~\cite{Ray2019} and SAC~\cite{ma2021feasible, yang2021wcsac}. 
Our PPO implementation is similar to~\cite{Ray2019} except that our multiplicative value function adds a secondary mechanism that improves the learning stability. For our SAC integration, the gradient of the multiplicative value function already naturally results in something similar to a Lagrange multiplier.
Commonly, the safety constraint is formulated as a constraint budget over the expected safety cost, e.g.,~\cite{lagrange_new, Ray2019, ma2021feasible, peng2021lagrangepi, lagrange_pid, safety_transfer_learning}. However, only considering the expected safety at each time step can cause the realized cost to exceed the constraint budget~\cite{yang2021wcsac}. To provide better constraint satisfaction, worst-case analysis can be performed with the conditional value at risk~\cite{lagrange_new, yang2021wcsac}. 
We tackle this issue by using reachability analysis and imposing zero constraint violation probability in our experiments.
Another line of research improves the stability of the learning process by using derivatives and integrals of the constraint function yielding a PID approach~\cite{lagrange_pid,peng2021lagrangepi}. 
Similarly, our multiplicative value function improves stability by simplifying the learning task.
Lastly, there are works that ensure state-wise safety with a learned Lagrange multiplier~\cite{ma2021feasible} and introduce safety transfer learning~\cite{safety_transfer_learning}. 



\textbf{Primal approaches} solve the CMDP by computing a policy gradient that satisfies the constraints.
Prior works have achieved this in different ways, for example by searching the feasible policy in the trust region~\cite{cpo}, projecting the unconstrained policy~\cite{yang2020projection}, projecting the optimum of the constrained non-parametric policy optimization~\cite{FOCOPS}, restricting the policy via log-barrier functions~\cite{liu2020ipo}, alternating between objective improvement and constraint satisfaction~\cite{xu2021crpo} or deriving an equivalent unconstrained problem~\cite{zhang2022penalized, sootla2022saute}.
Since the primal algorithms are harder to implement but not superior in terms of performance, they are less popular than the primal-dual algorithms.


\textbf{Lyapunov approaches} address the CMDP by leveraging Lyapunov functions, which are used in control theory to prove the stability of a dynamical system. 
In terms of model-based RL, the Lyapunov function can be used to guarantee that an agent can be brought back to a region of attraction~\cite{model_based_rl_safe}. 
More recently, this approach has been applied to model-free RL~\cite{lyapunov_corl} and extended in~\cite{safe_reachability} by an exploratory policy that maximizes its knowledge about safety. 
In~\cite{Lypunov_Lagrange}, the policy optimization is constrained on the Lyapunov decrease condition, which is then relaxed with a Lagrange multiplier. 

\textbf{Intervention approaches} use backup policies to ensure safe actions. 
Wagener et al.~\cite{advantage_based_intervention} defines an intervention rule based on the safety advantage between the action proposed by the backup policy and the RL agent. 
Another possibility is to construct a safe set using model-based or learning-based approaches, or a combination of both. 
Examples are control barrier functions~\cite{control_barrier_functions, control_barrier_eth}, reachability methods~\cite{reach_eth, reach_sim2real} or predictive safe set algorithms~\cite{backup_other, wabersich2021predictive, wachi2020safe}.
  
\section{Preliminaries}
\label{sec:preliminaries}
\PAR{Lagrangian methods.}
Let us consider the optimization problem $\min_x f(x) , \text{ st. } g(x) \leq c$, using Lagrangian primal-dual methods this problem can be cast to an unconstrained problem
\begin{equation*}
    (x^*,\lambda^*) = \min_x \max_{\lambda \geq 0} f(x) + \lambda (g(x) - c) \,,
\end{equation*}
where $\lambda$ denotes the dual variable or Lagrange multiplier~\cite{lagrangian_methods_bertsekas}.

\PAR{CMDP formulation.}
The interaction between the agent and the environment can be modeled with a Markov Decision Process (MDP). The MDP is defined by the Tuple $(\mathcal{S}, \mathcal{A}, r, P, s_0)$. Here $\mathcal{S}$ and $\mathcal{A}$ denote the state and action space respectively, the transition probability is $P(\cdot|s,a)$ and the reward is $r(s,a)$. Finally, $s_0 \in \mathcal{S}$ is the initial state. For simplicity, we consider deterministic rewards and initial states, but our results can be easily generalized to random initial state distributions and rewards. Extending an MDP with constraints yields a CMDP which is described by the tuple $(\mathcal{S}, \mathcal{A}, r, P, s_0, r_c, c_{\max})$. Here, $r_c$ is a safety cost and $c_{\max} \in \mathbb{R}_{\geq 0}$ is an upper bound on the expected cumulative safety cost. This yields the safety constraint $ \mathbb{E}^\pi \left [ \sum_{t=0}^{\infty} \gamma_c^t r_c(s_t) \middle | s_0 \right ]\leq c_{\max}$. The objective of the CMDP is to find an optimal policy $\pi^*$ according to
\begin{equation}
\label{eq:cmdp}
\begin{aligned}
\max_{\pi \in \Delta} \quad &  \mathbb{E}^\pi \left[ 
    \sum_{t=0}^{\infty} \gamma^t r(s_t, a_t) \middle| s_0  
    \right]\,, \\ 
\textrm{s.t.} \quad & \mathbb{E}^\pi \left [ \sum_{t=0}^{\infty} \gamma_c^t r_c(s_t) \middle | s_0 \right ]\leq c_{\max}\,.
\end{aligned}
\end{equation}
If the feasibility set induced by the safety constraint in Eq.~\ref{eq:cmdp} is non-empty, then there exists an optimal policy $\pi^*$ in the class of stationary Markovian policies $\Delta$~\cite{cmdp}.

\PAR{Reachability.}
To make the safety constraint in Eq.~\ref{eq:cmdp} more tangible, we specify the (un)safety as $P(\exists k : s_k \in \mathcal{C} |s_0=s_t)$, i.e., the probability of visiting states in the constraint set $\mathcal{C} \subseteq \mathcal{S}$. We consider constraint violations as catastrophic, thus states $s \in \mathcal{C}$ are terminal states. Coming back to the CMDP, the reachability problem can be cast to a value function by setting $r_c(s_t) = \mathds{1}_\mathcal{C}(s_t)$,
\begin{equation}
\label{eq:reachability}
\begin{aligned}
& P(\exists k : s_k \in \mathcal{C} |s_0=s_t) :=\Phi^{\pi}(s_t) \\
& =  \mathbb{E}^\pi \left [ \sum_{k=0}^{\infty} \gamma_c^k \mathds{1}_\mathcal{C}(s_k) \middle | s_0 = s_t\right ] \,,
\end{aligned}
\end{equation}
where $\mathds{1}_\mathcal{C}$ is the indicator function of the constraint set $\mathcal{C}$.

For the proof, we closely follow~\cite{lagrange_old}. First, we note that the sum $R_c :=  \sum_{k=0}^{\infty} \gamma_c^k r_c(s_k) $ is finite and at most $1$, namely when a constraint violation occurs and we reach a terminal state. Thus, when setting $\gamma_c=1$, it holds that $R_c = 1 \text{ if } \exists t : s_t \in C \text{ else } 0$. We note that $R_c$ is a Bernoulli Random variable and define $P(R_c=1):=q$. From the Bernoulli distribution we know that  $\mathbb{E}[R_c] = q =\Phi^\pi(s_t)$.  

Practically, a lower discount factor can increase the learning stability when used in an RL setup~\cite{lagrange_old}. Furthermore, we denote $\Phi^{\pi}(s_t)$ as the safety critic and similar to Eq.~\ref{eq:reachability}, we define the action value safety critic as $\Psi^\pi(s_t,a_t)$.

\section{Methods}
\label{sec:methods}
\PAR{Environment structure.}
We assume the following bounded reward structure of the environment
\begin{align} \label{eq:unclipped_reward}
    r(s_t,a_t) =\begin{cases}
      r_\textrm{constraint} & \textrm{if } s_t \in \mathcal{C} \\
      r_\textrm{constraint\_free}(s_t ,a_t) & \textrm{else}\\
    \end{cases} \,,
\end{align}
with $r_\textrm{constraint} \ll \min_{s,a}r_\textrm{constraint\_free}(s,a)$ and the constraint set $\mathcal{C}$ being a terminal state. The low reward for violating the constraint discourages standard RL agents from executing constraint violating actions. However, as shown in Fig.~\ref{fig:f_motivation}, the difference of magnitude between $r_\textrm{constraint}$ and $r_\textrm{constraint\_free}$ can cause discontinuities in the value landscape that are difficult to learn.

\PAR{Multiplicative value function.}
The motivation behind our multiplicative value function is to facilitate the learning by splitting responsibilities. The safety critic $\Phi^{\pi}(s_t)$ explicitly handles constraints, whereas the reward critic $\bar{V}^\pi(s_t)$ only estimates constraint-free returns. As argued in Sec.~\ref{sec:intro}, it is neither favorable to specify a magnitude for $r_\text{constraint}$ nor to learn $r_\text{constraint}$ with the reward critic $\bar{V}^\pi(s_t)$. Instead, we propose to clip the reward in Eq.~\ref{eq:unclipped_reward}, and learn the reward critic with this constrain neglecting reward,
\begin{align*}
& \bar{r}(s_t,a_t)=\begin{cases}
      \min_{s,a} r_\textrm{constraint\_free}(s,a) & \textrm{if } s_t \in \mathcal{C} \\
      r_\textrm{constraint\_free}(s_t ,a_t) & \textrm{else}\\
    \end{cases} \,,\\
 & \bar{V}^\pi(s_t) = \mathbb{E}^\pi \left[ \sum_{k=0}^{\infty} \gamma^k \bar{r}(s_k, a_k) \middle| s_0 = s_t \right]  \,.
\end{align*}
By taking the minimum in case of a constraint violation, we lightly discourage the policy from constraint violating actions. This can be especially useful when approximation errors cause the safety critic to be overly optimistic. Finally, the multiplicative value function 
$V_\text{mult}^\pi(s_t)$ 
is obtained by discounting the reward critic with the probability of constrained satisfaction:
\begin{align*}
V_\text{mult}^\pi(s_t) := 
 \left (\bar{V}^\pi(s_t) - \bar{v}_{\min}  \right ) \cdot \left (1 - \Phi^{\pi}(s_t)
\right ) + \bar{v}_{\min} \,,  
\end{align*}
where $\bar{v}_{\min}:=\min_{s} \bar{V}^\pi(s)$ is the lower bound on the reward critic, such that $\left (\bar{V}^\pi(s_t) - \bar{v}_{\min}  \right ) \geq 0$. Practically, we set $\bar{v}_{\min}$ to the minimum encountered $\bar{V}^\pi$ value during training. The multiplicative combination of the two critics allows a hyperparameter-free fusion, 
where a constraint violating state is associated with the value of $\bar{v}_{\min}$ and for save states, the value is $\bar{V}^\pi(s_t)$.
Similarly, we define $Q_\text{mult}^\pi(s_t,a_t) $ with $\bar{q}_{\min}$ as the multiplicative action value function: 
\begin{equation}
\label{eq:q_mult}
\begin{aligned}
& Q_\text{mult}^\pi(s_t,a_t) \\ 
& := 
 \left [
 \bar{Q}^\pi(s_t,a_t)  - \bar{q}_{\min} \right ] \cdot (1 - \Psi^\pi(s_t,a_t)) + \bar{q}_{\min} 
 \,.
\end{aligned}
\end{equation}
Note that the offset terms $\bar{v}_{\min}$ and $\bar{q}_{\min}$ could be avoided by assuming positive rewards. Nevertheless, introducing this term allows our formulation to handle arbitrary reward functions.

\PAR{Multiplicative advantage.}
We also want to consider advantage-based policy gradient methods. The advantage $A^\pi(s_t,a_t)$ is defined as,
\begin{equation}
\label{eq:advantage}
A^\pi=Q^\pi - V^\pi
= \left [ r(s_t,a_t) + \gamma V^\pi(s_{t+1}) \right ] - V^\pi(s_{t}) \,. 
\end{equation}
From this, we derive three versions of the multiplicative advantage $A_\text{mult}^\pi$:
\begin{align} 
& \text{V1: }  \left [
 \bar{r}_t  + \gamma V_\text{mult}^\pi(s_{t+1})
 \right ]  - 
 V_\text{mult}^\pi(s_{t}) \nonumber \\
& \text{V2: }  Q_\text{mult}^\pi(s_t,a_t) - V_\text{mult}^\pi(s_t) \label{eq:V1} \\ 
& \text{V3: }  \left [ \bar{Q}^\pi(s_t,a_t)  - \bar{q}_{\min} \right ] 
\left [1 - (r_{c,t} + \gamma_c \Phi^\pi(s_{t + 1})) \right ] \nonumber\\
& \text{ \color{white}{V3: }  } + \bar{q}_{\min}  -V_\text{mult}^\pi(s_t) \nonumber
\end{align}
In V1 and V2, we consider Eq.~\ref{eq:advantage} and replace all value functions with their multiplicative counterparts. 
Finally, V3 is similar to V2, but in Eq.~\ref{eq:q_mult} we use temporal difference bootstrapping for the safety critic.
\PAR{Integration into SAC.}
We integrate the multiplicative value function $Q^\pi_\text{mult}$ into the actor objective of SAC by replacing  $Q^\pi$ with $Q_\text{mult}^{\pi}$,
\begin{equation}
\label{eq:sac_objective_actor}
\max_{\theta} \mathbb{E}_{a_\theta \sim \pi_\theta} [Q_\text{mult}^{\pi_\theta}(s,a_\theta) - \alpha \log \pi_\theta(a_\theta|s)] \,.
\end{equation}
We call this version SAC Mult. For a compact notation, we drop the dependency on $(s,a_\theta, \pi)$ in the following. To get a better intuition about Eq.~\ref{eq:sac_objective_actor}, we investigate the gradient of the SAC Mult objective 
\begin{equation*}
\nabla_\theta Q_\text{mult}= (1  - \Psi) \cdot \nabla_\theta \bar{Q} -\left ( \bar{Q} -\bar{q}_{\min} \right ) \cdot \nabla_\theta \Psi ,
\end{equation*}
which has two terms. The first term is the gradient of the $\bar{Q}$-function discounted by the probability of constraint satisfaction. The second term is the gradient of the safety critic $\nabla_\theta \Psi$ discounted by $\left ( \bar{Q} -\bar{q}_{\min} \right )$, which can be understood as q-weighted multiplier. The disadvantage of this formulation is that in states where $\bar{Q}$ is high, the q-weighted multiplier becomes large and the gradient of the safety critic dominates the overall gradient. This can yield overly conservative behaviors. To mitigate this issue, we additionally propose two heuristics, SAC Mult Clipped
\begin{equation*}
    \nabla_\theta Q_\text{mult} \approx (1  - \Psi) \cdot \nabla_\theta \bar{Q}
    - \min \left ( \bar{Q} -\bar{q}_{\min}, \lambda_{\max} \right ) \cdot \nabla_\theta \Psi
\end{equation*}
and SAC Mult Lagrange
\begin{equation*}
    \nabla_\theta Q_\text{mult} \approx (1  - \Psi) \cdot \nabla_\theta \bar{Q}
    -\lambda \cdot \nabla_\theta \Psi\,.
\end{equation*}
In Mult Clipped, we limit the magnitude of the multiplier with $\lambda_{\max}$ which is a hyperparameter. In Mult Lagrange, we replace the q-weighted multiplier with a Lagrange multiplier that is optimized using primal-dual optimization. Under mild assumptions, this is guaranteed to converge to a local optimum of the CMDP~\cite{lagrange_new}. The Mult Lagrange objective is $\max_{\theta} \min_{\lambda \geq 0} \mathbb{E}_{a_\theta \sim \pi_\theta} [(1  - \Psi(s, a_\theta))\textrm{.detach()} \cdot \bar{Q}(s, a_{\theta}) - \alpha \log \pi_\theta(a_\theta|s) - \lambda \cdot (\Psi(s, a_\theta)-c_{\max})] \,,$
where .detach() denotes the operation of detaching the variable from the computational graph. For the exact implementation, we refer to our code.

\PAR{Integration into PPO.}
Given the three versions of the multiplicative advantage $A_\text{mult}^\pi$ in Eq.~\ref{eq:V1}, we can integrate them into the actor objective of PPO and extend it with a Lagrange multiplier
\begin{align*}
  \max_{\theta} \min_{\lambda \geq 0} \mathbb{E}_{s,a \sim \pi_{\theta_k}} [
\min \left \{  \frac{\pi_\theta(a|s)}{\pi_{\theta_k}(a|s)} A_\text{mult}^{\pi_{\theta_k}} , \text{   } g(\epsilon,  
A_\text{mult}^{\pi_{\theta_k}} )  \right \} 
\\
- \lambda \cdot \mathbb{E}_{a_\theta \sim \pi_\theta}[\Psi^\pi(s, a_\theta) - c_{\max}]
]  ,
\end{align*}
with $g(\epsilon, A) = (1 + \epsilon) A \cdot \mathbf{1}_{A \geq 0} + (1 - \epsilon) A \cdot \mathbf{1} _{A < 0}$. For the optimization of the Lagrange multiplier, we again proceed as in~\cite{lagrange_new} to guarantee convergence to a locally optimal policy. 
 

\section{Experimental Results}
\label{sec:result}

\begin{table}
\setlength{\tabcolsep}{3.3pt}
\centering
\caption{SAC and PPO evaluation results. Overall, PPO Mult V1 delivers most consistently good performance across environments. 
This might be because V1 is based on the Generalized Advantage Estimation~\cite{gae} which has shown to work particularly well for standard PPO. Among the SAC derivatives, Mult Clipped and Mult Lagrange perform the most consistent. For SAC Mult, the q-weighted safety multiplier yields overly conservative behavior in Lunar Lander, where the agent rarely lands, but instead times out. Consequently the reward is low.}
\label{tab:sac_and_ppo}
\begin{tabular}{lcccc} 
\toprule
& \begin{tabular}{@{}c@{}} Reward \\  $\uparrow$ \end{tabular} & \begin{tabular}{@{}c@{}} \% Constraint \\ violations $\downarrow$ \end{tabular}  & \begin{tabular}{@{}c@{}} Reward \\  $\uparrow$ \end{tabular} & \begin{tabular}{@{}c@{}} \% Constraint \\ violations $\downarrow$ \end{tabular}  \\
\midrule
& \multicolumn{4}{c}{Lunar Lander Safe} \\
\cmidrule(lr){2-5}
SAC & \multicolumn{2}{c}{50k} & \multicolumn{2}{c}{150k} \\
\cmidrule(lr){1-1}\cmidrule(lr){2-3}\cmidrule(lr){4-5}
SAC base & 90 $\pm$ 108 & 10 $\pm$ 16 & 181 $\pm$ 117 & 3 $\pm$ 6 \\ 

Lagrange & $111 \pm 105$ & $17 \pm 13$ & $184 \pm 128$ & $2 \pm 3$  \\

Mult & $-35 \pm 27$ & \textbf{3 $\pm$ 5} & $-34 \pm 22$ & $ 3 \pm 4$  \\

Mult Clipped & \textbf{134 $\pm$ 94} & $14 \pm 13$ & $243 \pm 49$ & $8 \pm 15$  \\

Mult Lagrange & $125 \pm 59$ & $29 \pm 15$ & \textbf{251 $\pm$ 20} & \textbf{2 $\pm$ 2}  \\
\cmidrule(lr){1-1}\cmidrule(lr){2-3}\cmidrule(lr){4-5}
PPO & \multicolumn{2}{c}{50k} & \multicolumn{2}{c}{150k} \\
\cmidrule(lr){1-1}\cmidrule(lr){2-3}\cmidrule(lr){4-5}
PPO base & $-126 \pm 158 $& 77 $\pm$ 29 & 225 $\pm$ 100 & 10 $\pm$ 30 \\ 

Lagrange & $-24 \pm 146$ & $54 \pm 39$ & $204 \pm 116$ & $12 \pm 24$ \\

V1 & $101 \pm 84$ & $41 \pm 19$ & $205 \pm 78$ & $7 \pm 16$ \\

V2 & $89 \pm 122$ & $44 \pm 34$ & $251 \pm 28$ & $5 \pm 9$ \\

V3 & \textbf{144 $\pm$ 4} & \textbf{26 $\pm$ 22} & \textbf{264 $\pm$ 5} & \textbf{1 $\pm$ 2} \\

FOCOPS & $-129 \pm 21$ & $64 \pm 24$ & $117 \pm 80$ & $30 \pm 19$ \\

\midrule
& \multicolumn{4}{c}{Point Robot Navigation} \\
\cmidrule(lr){2-5}
SAC & \multicolumn{2}{c}{50k} & \multicolumn{2}{c}{100k} \\

\cmidrule(lr){1-1}\cmidrule(lr){2-3}\cmidrule(lr){4-5}
SAC base & $22 \pm 19$ & 1 $\pm$ 1 & \textbf{38 $\pm$ 1} & 1 $\pm$ 1 \\ 

Lagrange & $29 \pm 8$ & \textbf{0 $\pm$ 0} & \textbf{38 $\pm$ 1} & \textbf{0 $\pm$ 0} \\

Mult & $34 \pm 3$ & \textbf{0 $\pm$ 0} & \textbf{38 $\pm$ 1} & \textbf{0 $\pm$ 0} \\

Mult Clipped & $33 \pm 4$ & \textbf{0 $\pm$ 0} & \textbf{38 $\pm$ 1} & \textbf{0 $\pm$ 0}  \\

Mult Lagrange & \textbf{37 $\pm$ 2} & \textbf{0 $\pm$ 0} & $37 \pm 2$ & \textbf{0 $\pm$ 0} \\
\cmidrule(lr){1-1}\cmidrule(lr){2-3}\cmidrule(lr){4-5}
PPO & \multicolumn{2}{c}{250k} & \multicolumn{2}{c}{500k} \\

\cmidrule(lr){1-1}\cmidrule(lr){2-3}\cmidrule(lr){4-5}
PPO base & 15 $\pm$ 15 & 3 $\pm$ 3 & 31 $\pm$ 3 & 3 $\pm$ 2\\ 

Lagrange & $15 \pm 16$ & 3 $\pm$ 3 & $24 \pm 5$ & $3 \pm 2$ \\

V1 & $25 \pm 5$ & 3 $\pm$ 3 & 30 $\pm$ 4 & 2 $\pm$ 1 \\

V2 & $11 \pm 21$ & 3 $\pm$ 3 & $17 \pm 15$ & $3 \pm 1$ \\

V3 & \textbf{27 $\pm$ 5} & $5 \pm 2$ & $29 \pm 3$ & $3 \pm 3$ \\

FOCOPS & $17 \pm 11$ & \textbf{0 $\pm$ 0} & \textbf{33 $\pm$ 2} & \textbf{0 $\pm$ 0}\\

\midrule
& \multicolumn{4}{c}{Gazebo Gym} \\
\cmidrule(lr){2-5}
SAC & \multicolumn{2}{c}{100k} & \multicolumn{2}{c}{200k} \\

\cmidrule(lr){1-1}\cmidrule(lr){2-3}\cmidrule(lr){4-5}
SAC base & 34 $\pm$ 6 & 7 $\pm$ 9 & 35 $\pm$ 5 & 6 $\pm$ 8 \\ 

Lagrange & $30 \pm 7$ & $11 \pm 12$ & $35 \pm 6$ & $5 \pm 10$ \\

Mult & $34 \pm 11$ & $5 \pm 12$ & $35 \pm 6$ & \textbf{3 $\pm$ 7} \\

Mult Clipped & \textbf{36 $\pm$ 5} & \textbf{3 $\pm$ 8} & \textbf{36 $\pm$ 5} & 3 $\pm$ 8\\

Mult Lagrange & 36 $\pm$ 6 & 4 $\pm$ 9 & \textbf{36 $\pm$ 5} & 3 $\pm$ 8 \\

\cmidrule(lr){1-1}\cmidrule(lr){2-3}\cmidrule(lr){4-5}
PPO & \multicolumn{2}{c}{250k} & \multicolumn{2}{c}{750k} \\
\cmidrule(lr){1-1}\cmidrule(lr){2-3}\cmidrule(lr){4-5}
PPO base & $27 \pm 6$ & $18 \pm 11$ & \textbf{31 $\pm$ 5} & $12 \pm 8$  \\

Lagrange & $26 \pm 5$ & $19 \pm 9$ & $31 \pm 6$ & $12 \pm 9$\\

V1 & \textbf{30 $\pm$ 6} & \textbf{14 $\pm$ 9} & \textbf{31 $\pm$ 5} & \textbf{11 $\pm$ 8} \\

FOCOPS & $13 \pm 12$ & $19 \pm 8$ & $29 \pm 5$ & $15 \pm 9$\\

\midrule
& \multicolumn{4}{c}{Car Racing Safe} \\
\cmidrule(lr){2-5}
PPO & \multicolumn{2}{c}{500k} & \multicolumn{2}{c}{1000k} \\

\cmidrule(lr){1-1}\cmidrule(lr){2-3}\cmidrule(lr){4-5}
PPO base & $43 \pm 23$ & $28 \pm 25$ & $89 \pm 25$ & $9 \pm 15$ \\

Lagrange & $43 \pm 23$ & $28 \pm 25$ & $89 \pm 25$ & $9 \pm 15$ \\

V1 & \textbf{74 $\pm$ 19} & \textbf{17 $\pm$ 14} & $98 \pm 13$ & \textbf{9 $\pm$ 7} \\

V2 & $65 \pm 24$ & $26 \pm 15$ & \textbf{100 $\pm$ 9} & $10 \pm 6$ \\

V3 & $73 \pm 21$ & $25 \pm 15$ & $78 \pm 17$ & $22 \pm 16$ \\

FOCOPS & $-2 \pm 2$ & $93 \pm 2$ & $-2 \pm 2$ & $93 \pm 2$\\

\bottomrule
\end{tabular}
\end{table}

We evaluate our methods in four environments: Lunar Lander Safe, Car Racing Safe, Point Robot Navigation, and Gazebo Gym. The first two are derived from OpenAI's gym \cite{openai_gym} and extended with safety constraints. For Lunar Lander Safe, we impose the constraint that the agent can only land within the landing zone. In Car Racing Safe, we test how our algorithm deals with both hard and soft constraints. We set the hard constraint that the agent is not allowed to leave the track. For the soft constraint, we encourage the agent to drive below 50 km/h. If the soft constraint is violated, the agent gets a small negative reward, but the episode still continues. The agent observes the environment via an agent centered bird's-eye-view image and a vector containing information about the steering angle, yawing rate, and velocity.
The last two environments focus on robotic navigation of ground robots. In Point Robot Navigation, the agent is a point robot that has to navigate towards the goal while avoiding obstacles. At each iteration, a new set of random obstacles is spawned and the agent starts at a random position. The agent perceives its environment via a local occupancy grid and the vector from the current position to the goal. The Gazebo Gym is similar but the agent is a Jackal differential drive robot modeled in Gazebo~\cite{gazebo} and the occupancy grid is replaced by a 1D-Lidar scan. Again, the task is to navigate to the goal that lies in the middle of a cluttered room. 

For the tuning, we use a sequential grid search on the learning rate, entropy coefficient, the number of optimization steps and experiment with the Beta distribution for the policy. 
For FOCOPS, we additionally tune lambda, the batch size and the KL divergence target. Furthermore, we tune the KL target and the clip range for PPO and the training frequency for SAC. After the baseline tuning, we keep the same hyperparameters for our multiplicative versions and additionally tune the safety discount factor $y_c$ and the initial value of the Lagrange multiplier $\lambda_\text{init}$. 

\begin{figure}
     \centering
     \begin{subfigure}[b]{0.48\textwidth}
         \centering
         \includegraphics[width=\textwidth]{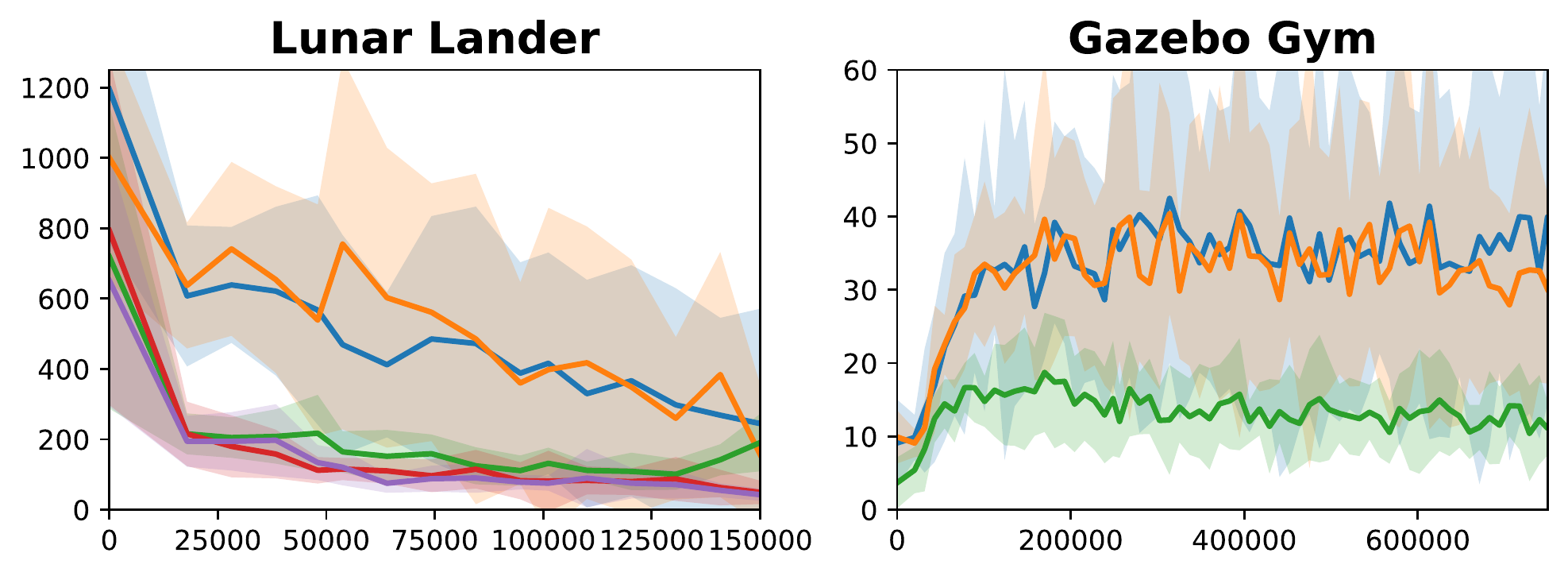}
         \caption{Value losses for PPO. Blue: Base, Orange: Lagrange, Green: Mult V1, Violet: Mult V2, Red: Mult V3.}
         \vspace{1ex}
     \end{subfigure}
     \begin{subfigure}[b]{0.48\textwidth}
         \centering
         \includegraphics[width=\textwidth]{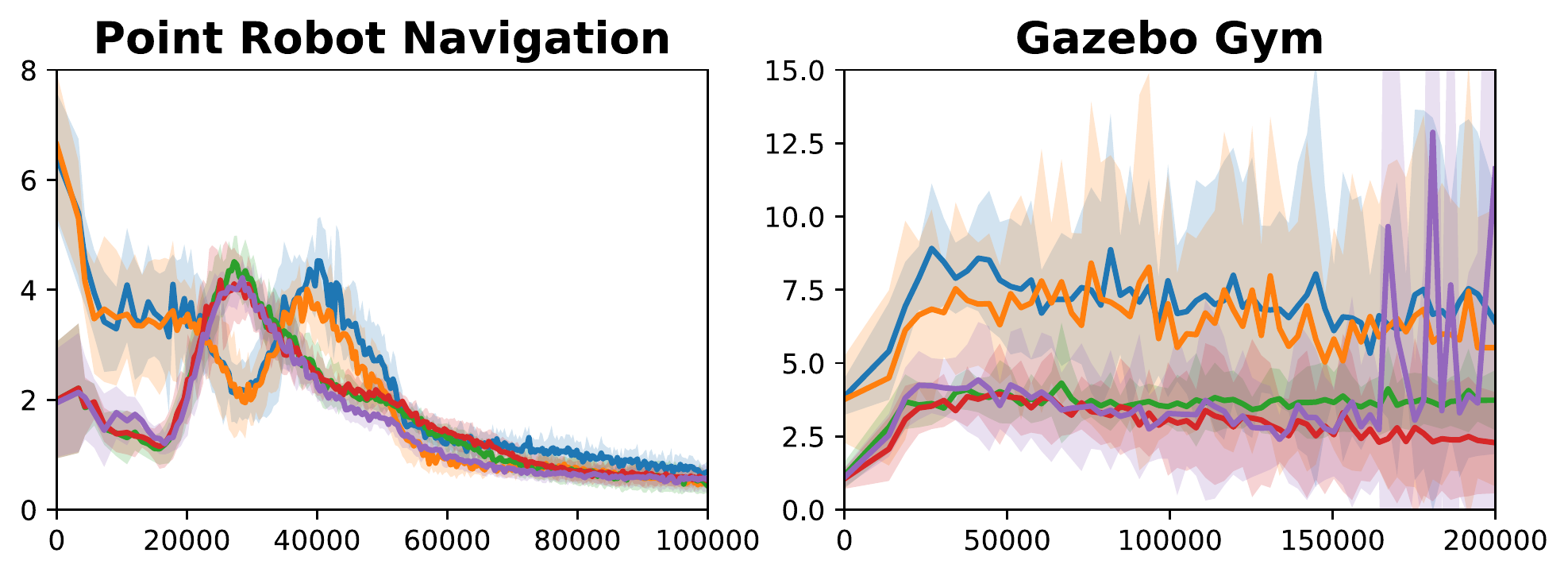}
         \caption{Value losses for SAC. Blue: Base, Orange: Lagrange, Green: Mult, Red: Mult Clipped, Violet: Mult Lagrange.}
         \vspace{1ex}
     \end{subfigure}

     \vspace{-1ex}
        \caption{Qualitative Results}
        \label{fig:loss}
        \vspace{-2ex}
\end{figure}

\subsection{Results and Comparisons}
With our experiments, we want to answer two questions:
Firstly, can the integration of the multiplicative value function facilitate the learning, leading to faster convergence and improved stability? For this, we integrate our approach 
into the SAC and PPO and compare against its Lagrangian counterpart.
Secondly, we ask, can the integration of the multiplicative value function into Lagrangian approaches yield comparable performance to recent approaches, e.g., FOCOPS~\cite{FOCOPS}? All the results are shown in Table~\ref{tab:sac_and_ppo}, where we evaluate each model at an intermediate checkpoint and at the end of the training. Each evaluation is over 10 seeds with 100 episodes.

\PAR{Increased sample efficiency.}
One of the main motivations for the multiplicative value function is to simplify the learning task. This is supported by Fig.~\ref{fig:loss}, where we observe a lower mean value loss and reduced variance across environments for both SAC and PPO. Having a simpler learning task allows our multiplicative versions to achieve significantly fewer constraint violations and higher rewards at the first evaluation checkpoint, indicating greater sampling efficiency. At the final evaluation checkpoint, our method achieves similar constraint satisfaction as the Lagrangian baseline. This is expected since both are Lagrangian methods and with enough training samples and model capacity, the regular value function can properly learn the value landscape. 

\PAR{Constraint satisfaction.}
In simpler environments, like Lunar Lander and Point Robot Navigation, our approach nearly achieves the target of zero constraint violations with PPO and SAC. In Car Racing Safe, the Lagrangian baseline and PPO V1 achieve 91\% constraint satisfaction. The imperfect performance could be caused by the challenging environment setup where minor driving errors can result in a crash. 
Due to the long run-time, we stopped the Gazebo Gym experiments in Table~\ref{tab:sac_and_ppo} before convergence.
In fact, SAC Mult Lagrange trained for 2 days as done for the sim-to-real transfer in Sec.~\ref{sec:real_world} achieves a constraint satisfaction of 100\%. 

\PAR{PPO vs. SAC.}
Overall, we achieve better constraint satisfaction with SAC agents in the navigation tasks Point Robot Navigation and Gazebo Gym. The only caveat is that we were not able to successfully train any SAC (nor FOCOPS) algorithm on Car Racing Safe because the agents never ``make" the first corner. Overall for PPO, we observe an increased variance in the training reward if the maximum allowed KL divergence is not explicitly tuned. The tuning is necessary because the multiplicative value function together with the Lagrange multiplier yields more aggressive policy updates which can cause instabilities.

\PAR{Soft constraint satisfaction.}
In Car Racing Safe, we additionally imposed the soft constraint to keep the velocity below 50 km/h. Even though PPO V1 and V2 achieve similar constraint satisfaction to the Lagrangian baseline, the reward is higher. This is because V1 and V2 violate the soft constraint with 20\% and 12\% respectively, whereas the Lagrangian Baseline violates the constraint in 32\% of the steps. We credit the multiplicative value function for the improved soft-constraint satisfaction. By facilitating the learning, the reward critic has more capacity to learn the fine details in the reward structure, like the soft constraint on the velocity.  

\PAR{Increased stability.}
We observe better training stability across seeds with the multiplicative value function. This is most prominent in Lunar Lander Safe, where we observe a large variance in the Lagrangian baseline rewards. This is because the Lagrangian agent only lands in 80\% of the seeds reliably, both for SAC and PPO. In contrast, SAC Mult Lagrange, Clipped and PPO Mult V2, V3 agents manage to land in all seeds, PPO Mult V1 in 90\% of the seeds. The poor performance of SAC Mult is not due to missing stability, in fact, SAC Mult performs badly across seeds. The issue is caused by the potentially large q-weighted multiplier, which makes the policy updates overly conservative leading to high timeout rates without ever landing.  

\begin{figure}
     \centering
     \begin{subfigure}[b]{0.45\textwidth}
         \centering
         \includegraphics[width=\textwidth]{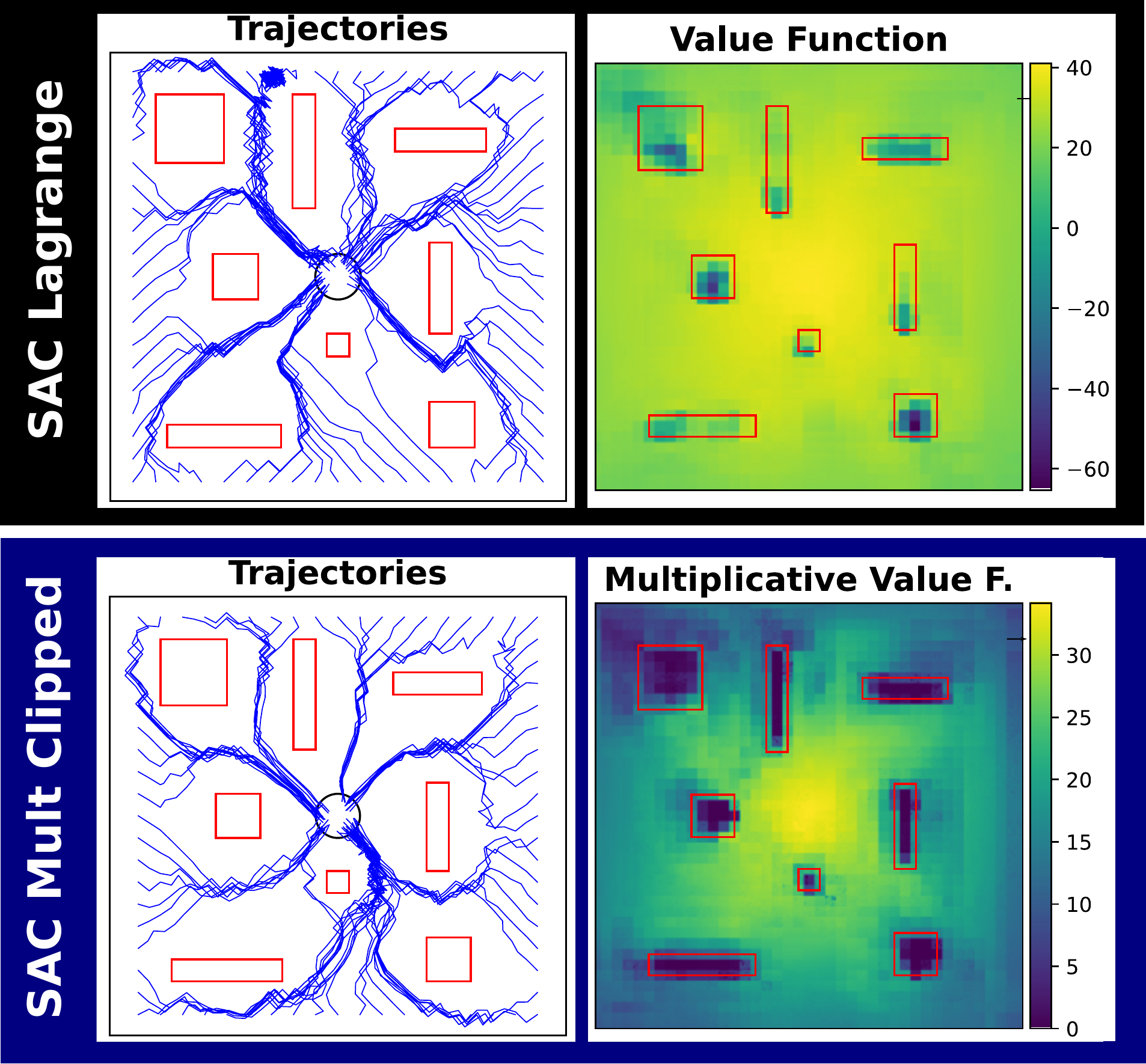}
         \caption{Evaluation of SAC Mult Lagrange and the Lagrangian baseline after 100k steps on Point Robot Navigation. The multiplicative value function better represents obstacles.}
         \vspace{1ex}
         \label{fig:boc_qualitative}
     \end{subfigure}
     \begin{subfigure}[b]{0.48\textwidth}
         \centering
         \includegraphics[width=\textwidth]{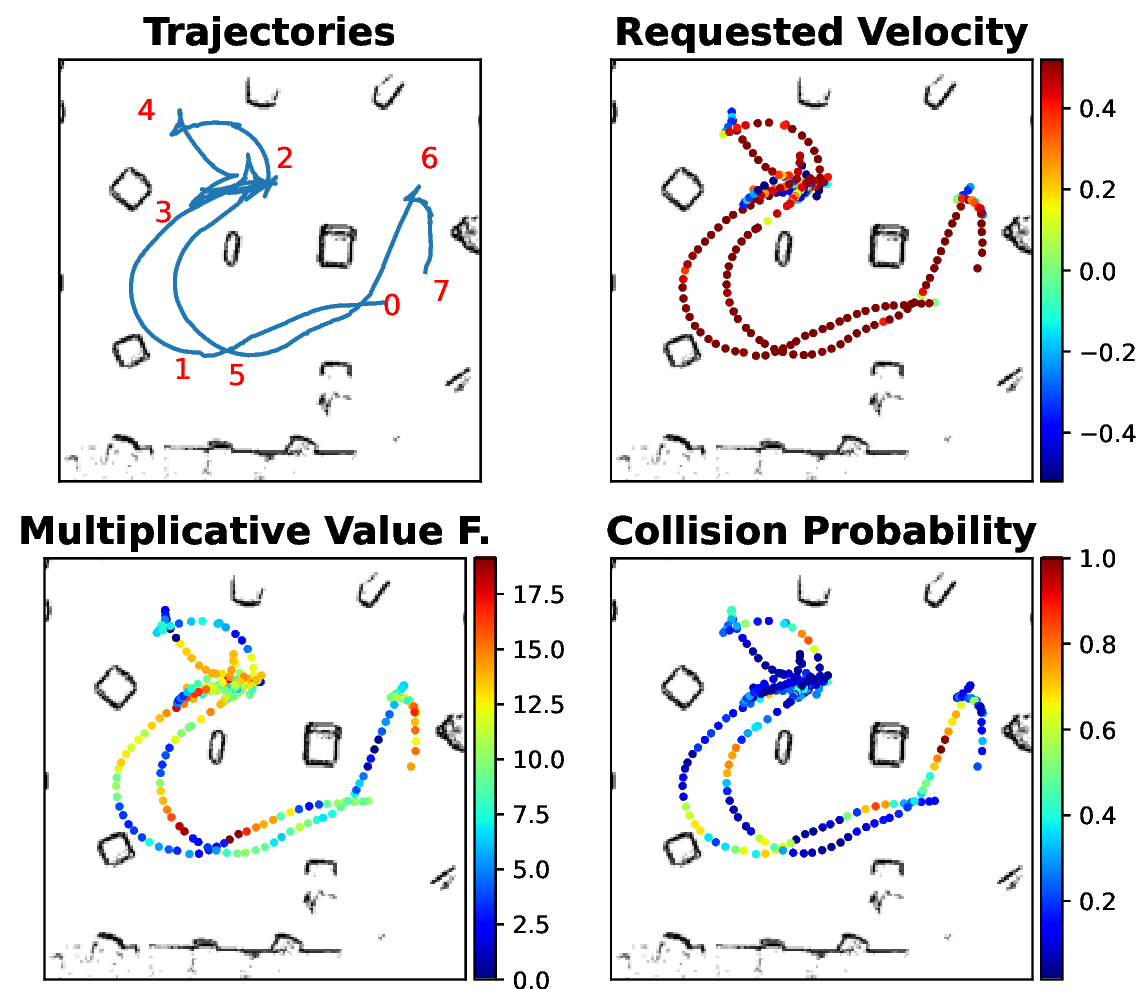}
         \caption{Trajectories of real-world experiment with a differential drive robot and SAC Mult Lagrange. Goal regions are marked with numbers 1-7. Starting point is 0. Here, success rate is 100\%.}
         \label{fig:real_world}
    \end{subfigure}
        \caption{Qualitative results on robot navigation environments.}
        \label{fig:qualitative_results}
        \vspace{-2ex}
\end{figure}

\PAR{Qualitative results.}
In Fig.~\ref{fig:boc_qualitative}, we show the qualitative results for Point Robot Navigation. Of most interest is the multiplicative value function, which can better represent the obstacles highlighted by red boxes. Furthermore, the trajectories of SAC Mult Clipped seem more directed towards the goal compared to the Lagrangian baseline. 
In Lunar Lander Safe, we observe the Mult agents land faster than the Lagrangian agents by having greater downward speeds high above the landing pad, while at lower altitudes, landing as cautiously as the Lagrangians.

\PAR{Theoretical guarantees vs. heuristics.}
Based on \cite{lagrange_new}, we have theoretical safety guarantees for all PPO Multiplicative versions as well as for SAC Mult Lagrange.
The SAC Mult and Clipped inhibit a q-weighted multiplier from the gradient of the multiplicative value function. However, this multiplier is not a Lagrangian multiplier, thus has no theoretical guarantees. Practically, we observe that SAC Mult and Mult Clipped have similar constraint satisfaction as SAC Mult Lagrange. 


\PAR{Comparison to FOCOPS.}
For Lunar Lander, the FOCOPS evaluation result at 150k steps is significantly worse than for any PPO Mult algorithm. We suspect this is caused by poor sample efficiency. Therefore, we train FOCOPS up to 300k steps where it obtains a reward of $215 \pm 65$ and a constraint violation rate of $13 \pm 18$\%. This is still worse than any Mult algorithm at 150k steps and is due to two seeds showing high constraint violation rates of 53\% and 38\%. In Point Robot Navigation, our algorithms converge faster but finally, FOCOPS outperforms all PPO Mult agents and is only beaten by SAC. In Gazebo Gym, FOCOPS performs worse than PPO in both evaluations, however, longer training could yield more comparable performance. Finally, in Car Racing Safe the FOCOPS agent never gets passed the first corner similar to SAC.   

\begin{figure}
\centering
    \includegraphics[width=0.48\textwidth]{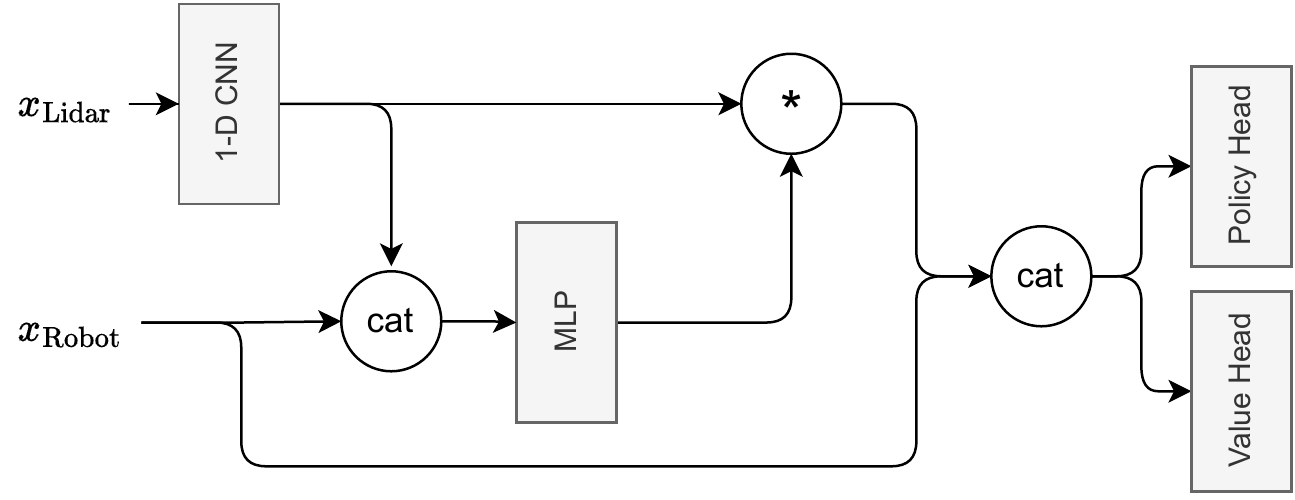}
    \caption{Attention Encoder used for the real world experiments. Here $x_\text{Robot}$ is the vector from current position to the goal and $x_\text{Lidar}$ is the measurement from the 1D-Lidar.}
    \label{fig:attention}
\end{figure}


\begin{figure*}
    \centerline{\includegraphics[width=\textwidth]{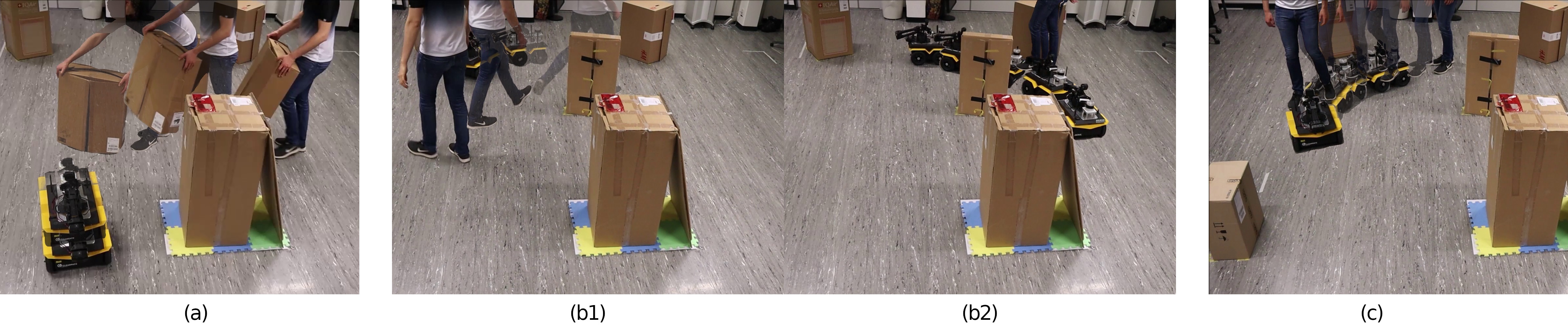}}
    \caption{(a): We dynamically put an obstacle in the trajectory of the robot which causes it to stop. The robot then waits until the obstacle is removed and continues its trajectory. 
    (b1, b2): Dynamic interactions with the robot by first walking next to it (b1), then overtaking it and standing to the side of the box (b2). While the person walks next to it the robot continues its trajectory. When the person overtook the robot, it swings to the right to avoid the collision. (c): The person walks towards the robot. The avoid the collision, the robot drives backwards. The complete video is available at \url{https://youtu.be/gAcETwOWTM4}.
    }
    \label{fig:all_overlay}
    \vspace{-2ex}
\end{figure*}


\subsection{Real-World Experiments}
\label{sec:real_world}
Based on the good performance of SAC Mult Lagrange on Gazebo Gym, the question arises of how the learned policy performs on a real Jackal robot? Can we tackle navigation, one of the fundamental robotic problems, in a safe way while only given sparse Lidar observations and a direction to the goal? To this end, we trained the policy for 4M steps in simulation with action noise, a smaller goal region of 0.3m and noisy Lidar observations. Furthermore, we included a cross-attention encoder for both the policy and value nets as depicted in Fig~\ref{fig:attention}. This attention mechanism allows the networks to focus on the latent representation of certain Lidar rays, for example the rays that are pointing forward. Those changes helped the policy to achieve a 100\% success rate with 0\% constraint violations in the simulation. 

One of the main differences between simulation and reality is that the ground friction in the real world is larger and the velocity controller behaves differently, i.e., given the same velocity command, the robot moves faster in simulation. In reality, if the robot is stationary and the velocity command is chosen too small, the robot can remain stationary due to the increased friction and different behavior of the velocity controller. Another real-world difficulty is the delay of 0.3s from the Lidar rays being recorded, send to the off-board computer, and the policy commands being sent back and executed on the robot. However, the fact that the real robot moves slower mitigates the time delay to an extent. 

In the lab, we constructed a cluttered obstacle course and directly deployed the policy trained in simulation on the robot. We defined seven goal regions and let the robot pass them four times. One of the four runs is depicted in Fig.~\ref{fig:real_world}. The robot drives with a direct trajectory from start 0 to goals 1 and 2, where it needs to reverse its direction by 180 degrees to approach goal 3. Interestingly, the policy did not learn to turn on the spot a differential drive robot would allow, but instead, the trajectories from goal 2-4 resemble more a kinematic bicycle model. This unfortunately caused the robot to get stuck once between goals 2-3, however, without violating a safety constraint. Overall, we report a success rate of 96\% with 100\% constraint satisfaction meaning that no box was touched. 

Encouraged by the demonstrated safety of our algorithm, we wanted to see if our agent can generalize from static box obstacles as encountered in the simulation to moving obstacles like humans in reality. For this, we arranged the goals in a circle and sequentially let the robot pass them. In the first experiment, we suddenly put a box in front of the agent as shown in Fig.~\ref{fig:all_overlay} (a). The robot reacted in a safe manner and came to an immediate stop. After a few seconds, we removed the box and the robot continued its original trajectory. 
In the second experiment shown in Fig.~\ref{fig:all_overlay} (b) we wanted to go a step further and see how the robot deals with obstacle shapes it has never seen before, i.e., human legs. Additionally, we wanted to know if the robot could naturally interact with a human walking next to it. For this, we started behind the robot, and then walked next to it at a certain safety distance, see Fig.~\ref{fig:all_overlay} (b1). This did not visibly influence the robot's trajectory. When we overtook the robot, we positioned ourselves close to a box such that we were in the trajectory of the robot, see Fig.~\ref{fig:all_overlay} (b2). Because of that, the robot corrected its trajectory and steered away from us. 
The final interaction is shown in Fig.~\ref{fig:all_overlay} (c). Here we wanted to investigate what happens if we actively provoke a collision by moving towards the robot. In that case, the robot started to move backward to keep a certain safety distance from us. The only drawback is that when relentlessly chasing the robot one can cause a rear collision with another obstacle. An explanation for this is that the robot has a Lidar blind spot in the back due to the mounted robot arm. All the interactions can be found in the supplementary video.  


\section{Conclusions and Limitations}
\label{sec:conclusion}

In this work, we introduced a safety critic to yield a multiplicative value function. We started with the CMDP formulation, derived the safety critic from reachability analysis and integrated our approach into the SAC and PPO framework. We proposed several versions of SAC and PPO using our multiplicative value function and showed increased sample efficiency and stability compared to the Lagrangian and FOCOPS baselines. Furthermore, the multiplicative value function can help to learn the fine details in the reward structure, like soft constraints. To show the real-world potential of our method, we took a SAC Mult Lagrange agent trained in simulation and successfully deployed the policy on a real robot in a zero-shot sim-to-real fashion. The robot showed safe behavior and was able to generalize to dynamic obstacles of novel shape. In future work, we would like to investigate further theoretical justification for our multiplicative value function.    

\PAR{Limitations.}
One of the main limitations is that our Lagrangian approach encourages safety during training but cannot guarantee it. 
This issue can be mitigated by adding an intervention mechanism, which could however cause problems when learning the safety critic as it requires reaching the constraint set. 
Future work will investigate the feasibility of using the intervention as a terminal state and more theoretical analysis of the multiplicative value function. Moreover, our method adds the initial value of the Lagrange multiplier and the safety discount factor $\gamma_c$ as hyperparameters to which the algorithm can be sensitive in some environments. 

   


    



\clearpage
\bibliographystyle{IEEEtran}
\bibliography{IEEEabrv,references.bib}

\begin{thebibliography}{10}
\providecommand{\url}[1]{#1}
\csname url@rmstyle\endcsname
\providecommand{\newblock}{\relax}
\providecommand{\bibinfo}[2]{#2}
\providecommand\BIBentrySTDinterwordspacing{\spaceskip=0pt\relax}
\providecommand\BIBentryALTinterwordstretchfactor{4}
\providecommand\BIBentryALTinterwordspacing{\spaceskip=\fontdimen2\font plus
\BIBentryALTinterwordstretchfactor\fontdimen3\font minus
  \fontdimen4\font\relax}
\providecommand\BIBforeignlanguage[2]{{%
\expandafter\ifx\csname l@#1\endcsname\relax
\typeout{** WARNING: IEEEtran.bst: No hyphenation pattern has been}%
\typeout{** loaded for the language `#1'. Using the pattern for}%
\typeout{** the default language instead.}%
\else
\language=\csname l@#1\endcsname
\fi
#2}}

\bibitem{atari}
V.~Mnih, K.~Kavukcuoglu, D.~Silver, A.~Graves, I.~Antonoglou, D.~Wierstra, and
  M.~Riedmiller, ``Playing atari with deep reinforcement learning,''
  \emph{arXiv}, 2013.

\bibitem{go}
D.~Silver, J.~Schrittwieser, K.~Simonyan, I.~Antonoglou, A.~Huang, A.~Guez,
  T.~Hubert, L.~Baker, M.~Lai, A.~Bolton, \emph{et~al.}, ``Mastering the game
  of go without human knowledge,'' \emph{nature}, vol. 550, no. 7676, pp.
  354--359, 2017.

\bibitem{starcraft}
O.~Vinyals, I.~Babuschkin, W.~M. Czarnecki, M.~Mathieu, A.~Dudzik, J.~Chung,
  D.~H. Choi, R.~Powell, T.~Ewalds, P.~Georgiev, \emph{et~al.}, ``Grandmaster
  level in starcraft ii using multi-agent reinforcement learning,''
  \emph{Nature}, vol. 575, no. 7782, pp. 350--354, 2019.

\bibitem{dota}
C.~Berner, G.~Brockman, B.~Chan, V.~Cheung, P.~Debiak, C.~Dennison, D.~Farhi,
  Q.~Fischer, S.~Hashme, C.~Hesse, \emph{et~al.}, ``Dota 2 with large scale
  deep reinforcement learning,'' \emph{arXiv}, 2019.

\bibitem{anymal}
T.~Miki, J.~Lee, J.~Hwangbo, L.~Wellhausen, V.~Koltun, and M.~Hutter,
  ``Learning robust perceptive locomotion for quadrupedal robots in the wild,''
  \emph{Science Robotics}, vol.~7, no.~62, p. eabk2822, 2022.

\bibitem{gran_turismo}
F.~Fuchs, Y.~Song, E.~Kaufmann, D.~Scaramuzza, and P.~D{\"u}rr, ``Super-human
  performance in gran turismo sport using deep reinforcement learning,''
  \emph{IEEE Robotics and Automation Letters}, vol.~6, no.~3, pp. 4257--4264,
  2021.

\bibitem{carla}
Z.~Zhang, A.~Liniger, D.~Dai, F.~Yu, and L.~Van~Gool, ``End-to-end urban
  driving by imitating a reinforcement learning coach,'' in \emph{Proceedings
  of the IEEE/CVF International Conference on Computer Vision}, 2021, pp.
  15\,222--15\,232.

\bibitem{cmdp}
E.~Altman, \emph{Constrained Markov Decision Processes}.\hskip 1em plus 0.5em
  minus 0.4em\relax CRC Press, 1999.

\bibitem{lagrange_old}
P.~Geibel and F.~Wysotzki, ``Risk-sensitive reinforcement learning applied to
  control under constraints,'' \emph{Journal of Artificial Intelligence
  Research}, vol.~24, pp. 81--108, 2005.

\bibitem{lagrange_new}
Y.~Chow, M.~Ghavamzadeh, L.~Janson, and M.~Pavone, ``Risk-constrained
  reinforcement learning with percentile risk criteria,'' \emph{The Journal of
  Machine Learning Research}, vol.~18, no.~1, pp. 6070--6120, 2017.

\bibitem{safety_transfer_learning}
K.~Srinivasan, B.~Eysenbach, S.~Ha, J.~Tan, and C.~Finn, ``Learning to be safe:
  Deep rl with a safety critic,'' \emph{arXiv}, 2020.

\bibitem{lagrange_pid}
A.~Stooke, J.~Achiam, and P.~Abbeel, ``Responsive safety in reinforcement
  learning by pid lagrangian methods,'' in \emph{ICML}.\hskip 1em plus 0.5em
  minus 0.4em\relax PMLR, 2020, pp. 9133--9143.

\bibitem{cpo}
J.~Achiam, D.~Held, A.~Tamar, and P.~Abbeel, ``Constrained policy
  optimization,'' in \emph{ICML}.\hskip 1em plus 0.5em minus 0.4em\relax PMLR,
  2017, pp. 22--31.

\bibitem{ppo}
J.~Schulman, F.~Wolski, P.~Dhariwal, A.~Radford, and O.~Klimov, ``Proximal
  policy optimization algorithms,'' \emph{arXiv}, 2017.

\bibitem{sac}
T.~Haarnoja, A.~Zhou, P.~Abbeel, and S.~Levine, ``Soft actor-critic: Off-policy
  maximum entropy deep reinforcement learning with a stochastic actor,'' in
  \emph{ICML}.\hskip 1em plus 0.5em minus 0.4em\relax PMLR, 2018, pp.
  1861--1870.

\bibitem{liu2021policy}
Y.~Liu, A.~Halev, and X.~Liu, ``Policy learning with constraints in model-free
  reinforcement learning: A survey.'' in \emph{IJCAI}, 2021, pp. 4508--4515.

\bibitem{gu2022review}
S.~Gu, L.~Yang, Y.~Du, G.~Chen, F.~Walter, J.~Wang, Y.~Yang, and A.~Knoll, ``A
  review of safe reinforcement learning: Methods, theory and applications,''
  \emph{arXiv}, 2022.

\bibitem{tessler2018reward}
C.~Tessler, D.~J. Mankowitz, and S.~Mannor, ``Reward constrained policy
  optimization,'' in \emph{Proceedings of the International Conference on
  Learning Representations (ICLR)}, 2019.

\bibitem{paternain2019constrained}
S.~Paternain, L.~Chamon, M.~Calvo-Fullana, and A.~Ribeiro, ``Constrained
  reinforcement learning has zero duality gap,'' \emph{Advances in Neural
  Information Processing Systems (NeurIPS)}, vol.~32, 2019.

\bibitem{Ray2019}
A.~Ray, J.~Achiam, and D.~Amodei, ``{Benchmarking Safe Exploration in Deep
  Reinforcement Learning},'' 2019.

\bibitem{ma2021feasible}
H.~Ma, Y.~Guan, S.~E. Li, X.~Zhang, S.~Zheng, and J.~Chen, ``Feasible
  actor-critic: Constrained reinforcement learning for ensuring statewise
  safety,'' \emph{arXiv preprint arXiv:2105.10682}, 2021.

\bibitem{yang2021wcsac}
Q.~Yang, T.~D. Sim{\~a}o, S.~H. Tindemans, and M.~T. Spaan, ``Wcsac: Worst-case
  soft actor critic for safety-constrained reinforcement learning.'' in
  \emph{AAAI}, 2021, pp. 10\,639--10\,646.

\bibitem{peng2021lagrangepi}
B.~Peng, Y.~Mu, J.~Duan, Y.~Guan, S.~E. Li, and J.~Chen, ``Separated
  proportional-integral lagrangian for chance constrained reinforcement
  learning,'' in \emph{2021 IEEE Intelligent Vehicles Symposium (IV)}.\hskip
  1em plus 0.5em minus 0.4em\relax IEEE, 2021, pp. 193--199.

\bibitem{yang2020projection}
T.-Y. Yang, J.~Rosca, K.~Narasimhan, and P.~J. Ramadge, ``Projection-based
  constrained policy optimization,'' in \emph{Proceedings of the International
  Conference on Learning Representations (ICLR)}, 2020.

\bibitem{FOCOPS}
Y.~Zhang, Q.~Vuong, and K.~Ross, ``First order constrained optimization in
  policy space,'' \emph{Advances in Neural Information Processing Systems
  (NeurIPS)}, vol.~33, pp. 15\,338--15\,349, 2020.

\bibitem{liu2020ipo}
Y.~Liu, J.~Ding, and X.~Liu, ``Ipo: Interior-point policy optimization under
  constraints,'' in \emph{AAAI}, vol.~34, no.~04, 2020, pp. 4940--4947.

\bibitem{xu2021crpo}
T.~Xu, Y.~Liang, and G.~Lan, ``Crpo: A new approach for safe reinforcement
  learning with convergence guarantee,'' in \emph{ICML}.\hskip 1em plus 0.5em
  minus 0.4em\relax PMLR, 2021, pp. 11\,480--11\,491.

\bibitem{zhang2022penalized}
L.~Zhang, L.~Shen, L.~Yang, S.~Chen, X.~Wang, B.~Yuan, and D.~Tao, ``Penalized
  proximal policy optimization for safe reinforcement learning,'' in
  \emph{IJCAI}, 2022.

\bibitem{sootla2022saute}
A.~Sootla, A.~I. Cowen-Rivers, T.~Jafferjee, Z.~Wang, D.~H. Mguni, J.~Wang, and
  H.~Ammar, ``Saut{\'e} rl: Almost surely safe reinforcement learning using
  state augmentation,'' in \emph{ICML}.\hskip 1em plus 0.5em minus 0.4em\relax
  PMLR, 2022, pp. 20\,423--20\,443.

\bibitem{model_based_rl_safe}
F.~Berkenkamp, M.~Turchetta, A.~Schoellig, and A.~Krause, ``Safe model-based
  reinforcement learning with stability guarantees,'' \emph{Advances in Neural
  Information Processing Systems (NeurIPS)}, vol.~30, 2017.

\bibitem{lyapunov_corl}
Y.~Chow, O.~Nachum, A.~Faust, E.~Duenez-Guzman, and M.~Ghavamzadeh, ``Safe
  policy learning for continuous control,'' in \emph{Conference on Robot
  Learning}, 2020.

\bibitem{safe_reachability}
S.~Huh and I.~Yang, ``Safe reinforcement learning for probabilistic
  reachability and safety specifications: A lyapunov-based approach,''
  \emph{arXiv}, 2020.

\bibitem{Lypunov_Lagrange}
L.~Zhang, R.~Zhang, T.~Wu, R.~Weng, M.~Han, and Y.~Zhao, ``Safe reinforcement
  learning with stability guarantee for motion planning of autonomous
  vehicles,'' \emph{IEEE Transactions on Neural Networks and Learning Systems},
  vol.~32, no.~12, pp. 5435--5444, 2021.

\bibitem{advantage_based_intervention}
N.~C. Wagener, B.~Boots, and C.-A. Cheng, ``Safe reinforcement learning using
  advantage-based intervention,'' in \emph{ICML}.\hskip 1em plus 0.5em minus
  0.4em\relax PMLR, 2021, pp. 10\,630--10\,640.

\bibitem{control_barrier_functions}
R.~Cheng, G.~Orosz, R.~M. Murray, and J.~W. Burdick, ``End-to-end safe
  reinforcement learning through barrier functions for safety-critical
  continuous control tasks,'' in \emph{AAAI}, vol.~33, no.~01, 2019, pp.
  3387--3395.

\bibitem{control_barrier_eth}
K.~P. Wabersich and M.~N. Zeilinger, ``Predictive control barrier functions:
  Enhanced safety mechanisms for learning-based control,'' \emph{IEEE
  Transactions on Automatic Control}, 2022.

\bibitem{reach_eth}
J.~F. Fisac, A.~K. Akametalu, M.~N. Zeilinger, S.~Kaynama, J.~Gillula, and
  C.~J. Tomlin, ``A general safety framework for learning-based control in
  uncertain robotic systems,'' \emph{IEEE Transactions on Automatic Control},
  vol.~64, no.~7, pp. 2737--2752, 2018.

\bibitem{reach_sim2real}
K.-C. Hsu, A.~Z. Ren, D.~P. Nguyen, A.~Majumdar, and J.~F. Fisac,
  ``Sim-to-lab-to-real: Safe reinforcement learning with shielding and
  generalization guarantees,'' \emph{arXiv}, 2022.

\bibitem{backup_other}
W.~Zhao, T.~He, and C.~Liu, ``Model-free safe control for zero-violation
  reinforcement learning,'' in \emph{Conference on Robot Learning}, 2021.

\bibitem{wabersich2021predictive}
K.~P. Wabersich and M.~N. Zeilinger, ``A predictive safety filter for
  learning-based control of constrained nonlinear dynamical systems,''
  \emph{Automatica}, vol. 129, p. 109597, 2021.

\bibitem{wachi2020safe}
A.~Wachi and Y.~Sui, ``Safe reinforcement learning in constrained markov
  decision processes,'' in \emph{ICML}.\hskip 1em plus 0.5em minus 0.4em\relax
  PMLR, 2020, pp. 9797--9806.

\bibitem{lagrangian_methods_bertsekas}
D.~P. Bertsekas, \emph{Constrained optimization and Lagrange multiplier
  methods}.\hskip 1em plus 0.5em minus 0.4em\relax Academic press, 2014.

\bibitem{gae}
J.~Schulman, P.~Moritz, S.~Levine, M.~Jordan, and P.~Abbeel, ``High-dimensional
  continuous control using generalized advantage estimation,'' in
  \emph{Proceedings of the International Conference on Learning Representations
  (ICLR)}, 2016.

\bibitem{openai_gym}
G.~Brockman, V.~Cheung, L.~Pettersson, J.~Schneider, J.~Schulman, J.~Tang, and
  W.~Zaremba, ``Openai gym,'' \emph{arXiv}, 2016.

\bibitem{gazebo}
N.~Koenig and A.~Howard, ``Design and use paradigms for gazebo, an open-source
  multi-robot simulator,'' in \emph{IEEE/RSJ International Conference on
  Intelligent Robots and Systems (IROS)}.\hskip 1em plus 0.5em minus
  0.4em\relax IEEE, 2004.

\bibitem{stable-baselines3}
A.~Raffin, A.~Hill, A.~Gleave, A.~Kanervisto, M.~Ernestus, and N.~Dormann,
  ``Stable-baselines3: Reliable reinforcement learning implementations,''
  \emph{Journal of Machine Learning Research}, 2021.

\bibitem{rl_zoo3}
A.~Raffin, ``Rl baselines3 zoo,''
  \url{https://github.com/DLR-RM/rl-baselines3-zoo}, 2020.

\bibitem{sde}
A.~Raffin, J.~Kober, and F.~Stulp, ``Smooth exploration for robotic
  reinforcement learning,'' in \emph{Conference on Robot Learning}.\hskip 1em
  plus 0.5em minus 0.4em\relax PMLR, 2022, pp. 1634--1644.

\end{thebibliography}

\appendix

\subsection{Supplementary Video}

The supplementary video for the paper can be found here at \url{https://youtu.be/gAcETwOWTM4}.
This video demonstrates the qualitative outcomes of the interaction experiment with the Jackal differential drive robot. In the first part of the video, we dynamically block the path of the agent with a box. Encouraged by the demonstrated safety, we try interactions with a human in the second part of the video. This is challenging in two ways: First, the agent only perceives the legs of the human, which are significantly thinner than the obstacles encountered in training. Second, the agent only encountered static objects in training while the human is a dynamic obstacle. Still, the agent shows safe behavior.

\subsection{Complete Algorithms for SAC and PPO Multiplicative}
Algorithm~\ref{alg:safe_sac} is the complete version of SAC Multiplicative, whereas Algorithm~\ref{alg:ppo} is the complete version of the PPO Multiplicative. The differences between our proposed methods and the standard algorithms are highlighted in blue.

\subsection{Hyperparameter Tuning}
For the tuning, we start with the default parameters of~\cite{stable-baselines3} or the suggested parameters of~\cite{rl_zoo3} for Car Racing and Lunar Lander. We then tune the PPO and SAC baseline algorithms by varying the parameters of the initial learning rate and schedule, entropy coefficient and schedule, state-dependent exploration~\cite{sde} vs. standard action sampling, and Gaussian vs. Beta distribution for the actor policy. For PPO, we additionally tune the clip range, the KL divergence target, the initial variance parameter of the Gaussian policy, and the number of epochs of optimization. As for SAC, we vary the training frequency, the number of gradient steps, how many samples to collect before starting the training, and the buffer size. Once satisfied with the baseline performance, we keep the same hyperparameters for our multiplicative versions and additionally tune the safety discount factor $y_c$ and the initial value of the Lagrange multiplier $\lambda_\text{init}$. To best compare the effect of the multiplicative value function, we use the same initial Lagrange multiplier $\lambda_\text{init}$ for our multiplicative versions and the Lagrange baseline.

\subsection{Detailed Experimental Description}
To recapitulate, we assume the following reward structure of the environment
\begin{align*}
    r(s_t,a_t) =\begin{cases}
      r_\textrm{constraint} & \textrm{if } s_t \in \mathcal{C} \\
      r_\textrm{constraint\_free}(s_t ,a_t) & \textrm{else}\\
    \end{cases} \,.
\end{align*}
Here, we associate $r_\textrm{constraint}$ with hard constraints which cause the episode to end in case of constraint violations. Additionally, there can be soft constraints encoded in $r_\textrm{constraint\_free}$. Violating a soft constraint causes a negative reward but the episode continues.

\textbf{Lunar Lander Safe} is a continuous control task where the agent has to land a rocket on the moon's surface~\cite{openai_gym}. Once the rocket lands, the episode ends. The agent receives a reward for minimizing the distance to the landing pad and it can land anywhere as long as its down velocity is slow enough when touching the ground. The observation is
\begin{equation*}
    \vec{x}_\text{observation} = [\vec{d}, \vec{v}, \phi, \dot{\phi}, \mathbbm{1}_\text{contact\_left}, \mathbbm{1}_\text{contact\_right}],
\end{equation*}
where $\vec{d}$ denotes the vector from current position to landing pad, $\vec{v}$ is the linear velocity of the agent, $\phi$ the roll angle and $\mathbbm{1}_\text{contact}$ denotes if the corresponding left or right leg has ground contact.
We want to make the environment more safety-critical and go by the concept ``the floor is lava". This means, we keep the original constraint on the maximum allowed landing velocity but expand the constraint set $\mathcal{C}$ such that landing outside the landing pad is not allowed anymore. The rewards are
\begin{equation*}
    \begin{aligned}
r_\textrm{constraint\_free} = & -100 \cdot \left(||\vec{d}_t||_2 - ||\vec{d}_{t-1}||_2 \right) \\
 & -100 \cdot \left(||\vec{v}_t||_2 - ||\vec{v}_{t-1}||_2 \right) \\
&-100 \cdot \left(\phi_t - \phi_{t-1} \right) \\
&+10 \cdot \left(  \mathbbm{1}_\text{contact\_left,t} - \mathbbm{1}_\text{contact\_left,t-1} \right)\\
&+10 \cdot \left(  \mathbbm{1}_\text{contact\_right,t} - \mathbbm{1}_\text{contact\_right,t-1} \right) \\
& - \text{fuel\_spend} \\
&+ 100 \cdot \mathbbm{1}_\text{contact\_right,t}  \mathbbm{1}_\text{contact\_left,t}  \mathbbm{1}_{||\vec{v}_t||_2 < 0.01} , \\ 
r_\textrm{constraint} = & -100,
\end{aligned}
\end{equation*}
where the last line in $r_\textrm{constraint\_free}$ denotes the state of a successful landing. Furthermore, we introduce a timeout if the agent cannot land within 1000 steps. Since no measure of time is present in the observation, the agent does not know about the timeout. 
The action space is two dimensional, representing the impulse the agent can apply to the left/right and up/down using ``rocket engines".

\textbf{Car Racing Safe} environment is based on CarRacing from OpenAI~\cite{openai_gym} where the agent is rewarded for driving around a race track. Each iteration, a new random track is spawned. The episode terminates if the agent has visited all track tiles or leaves the playground, which extends far beyond the track. Again, we want to make this environment more safety critic. For this, we tighten the hard constraint such that the agent enters the constraint set $\mathcal{C}$ if it leaves the racetrack. Also, the agent has to drive faster than 0.1 km/h after an initial time period. Additionally, we impose a soft constraint that encourages the agent to drive below 50 km/h, which  is encoded in the reward
\begin{equation*}
\begin{aligned}
r_\textrm{constraint\_free} = & 
\frac{0.3}{n_{tiles}} \cdot \mathbbm{1}_\text{new\_track}
+ \begin{cases}
0.005 v & \text{if } v < 50 \\
-0.01 v & \text{if } v \geq 50 \\
\end{cases}, \\
r_\textrm{constraint} = & -10 \,,
\end{aligned}
\end{equation*}
where $v$ denotes the longitudinal velocity. The observation space of the agent consists of an agent-centered bird's-eye-view image, longitudinal velocity $v$, yawing rate $\psi$ and steering angle of the wheels $\delta$,
\begin{equation*}
    \vec{x}_\text{observation} = [\text{img}, v, \dot{\psi}, \delta].
\end{equation*}

The action space is continuous and two-dimensional. The actions are between accelerating/braking and steering to the left/right. We use the CNN structure of~\cite{atari} to process the birds-eye-view image. 

\subsection{Point Robot Navigation}
\begin{figure}
    \centering
    \includegraphics[width=0.3\textwidth]{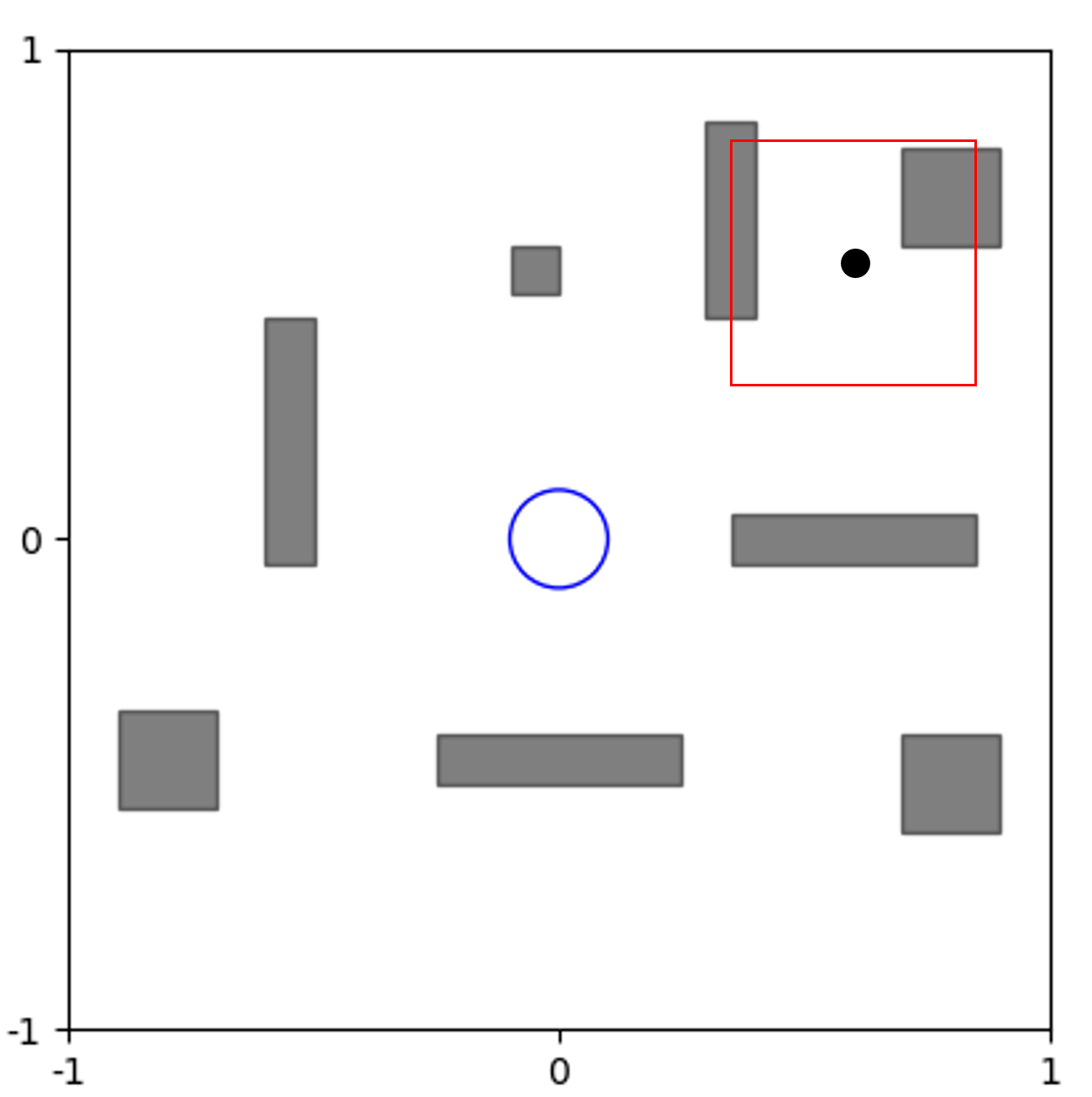}
    \caption{Point Robot Navigation. The obstacles are shown in gray, the goal region in blue, the agent is a black dot, and the field of view of the local occupancy grid is marked by the red box. The environment has the boundaries $[-1,1]$ m.}
    \label{fig:boc_env}
\end{figure}

In the Point Robot Navigation environment, the agent has to navigate to a goal region of 0.1 m while avoiding obstacles and staying inside the environment boundaries. At each iteration, both the starting position of the robot and the set of obstacles are random. An example of the environment can be taken from Fig.~\ref{fig:boc_env}. As the observation, the agent receives the vector $\vec{d}$ from the current position to the goal and an agent-centered occupancy grid,
\begin{equation*}
    \vec{x}_\text{observation} = [\vec{d}, \text{oc\_grid}].
\end{equation*}
The action space is two dimensional, representing the percentage of a step size the agent can travel in x and y-direction. The reward is
\begin{equation}
    \label{eq:boc_reward}
    \begin{aligned}
    r_\textrm{constraint\_free} = 
    & \begin{cases}
    40 & \text{if } ||\vec{d}||_2 < 0.1 \\
    -0.1 & \text{else} 
    \end{cases} \,, \\
    r_\textrm{constraint} = & -20 \,.
    \end{aligned}
\end{equation}
This corresponds to a sparse reward setting. The step penalty of $-0.1$ encourages the agent to reach the goal in a low number of steps. Furthermore, we choose the reward for reaching the goal higher than the penalty for constraint violation. With a lower goal reward, we have experienced baseline agents that try to crash immediately into obstacles to avoid the accumulation of step penalties. 
To process the occupancy grid, we use a small CNN encoder. When initially training SAC and PPO baselines, we experienced instability. This was related to the model not understanding the occupancy grid. To mitigate the issue, we added a decoder module to the CNN and posed an auxiliary loss on the reconstruction error. This improved the performance and stability of PPO. For SAC, we had to additionally use separate CNN encoders for actor, reward and safety critic.

\begin{figure}
     \centering
     \includegraphics[width=0.45\textwidth]{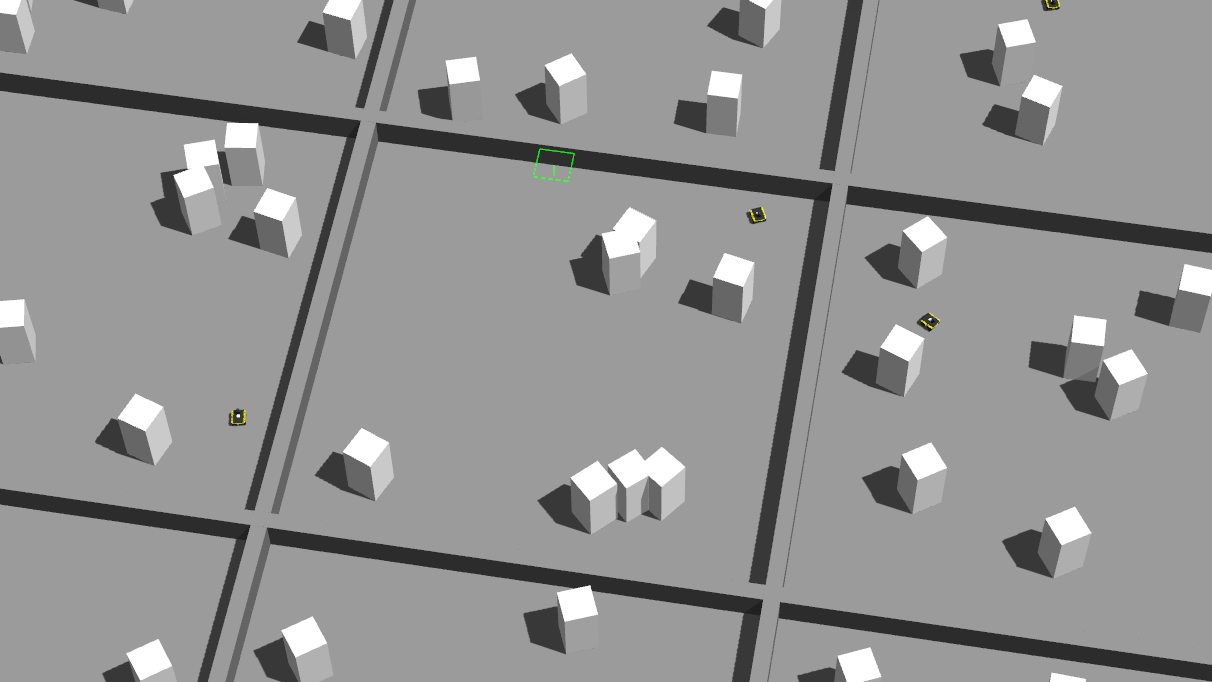}
     \caption{Gazebo Gym. This is an 8x8 m world where the task is to navigate a cluttered environment. The picture shows the training of PPO.}
     \label{fig:gg}
\end{figure}

\begin{figure}
     \centering
     \includegraphics[width=0.3\textwidth]{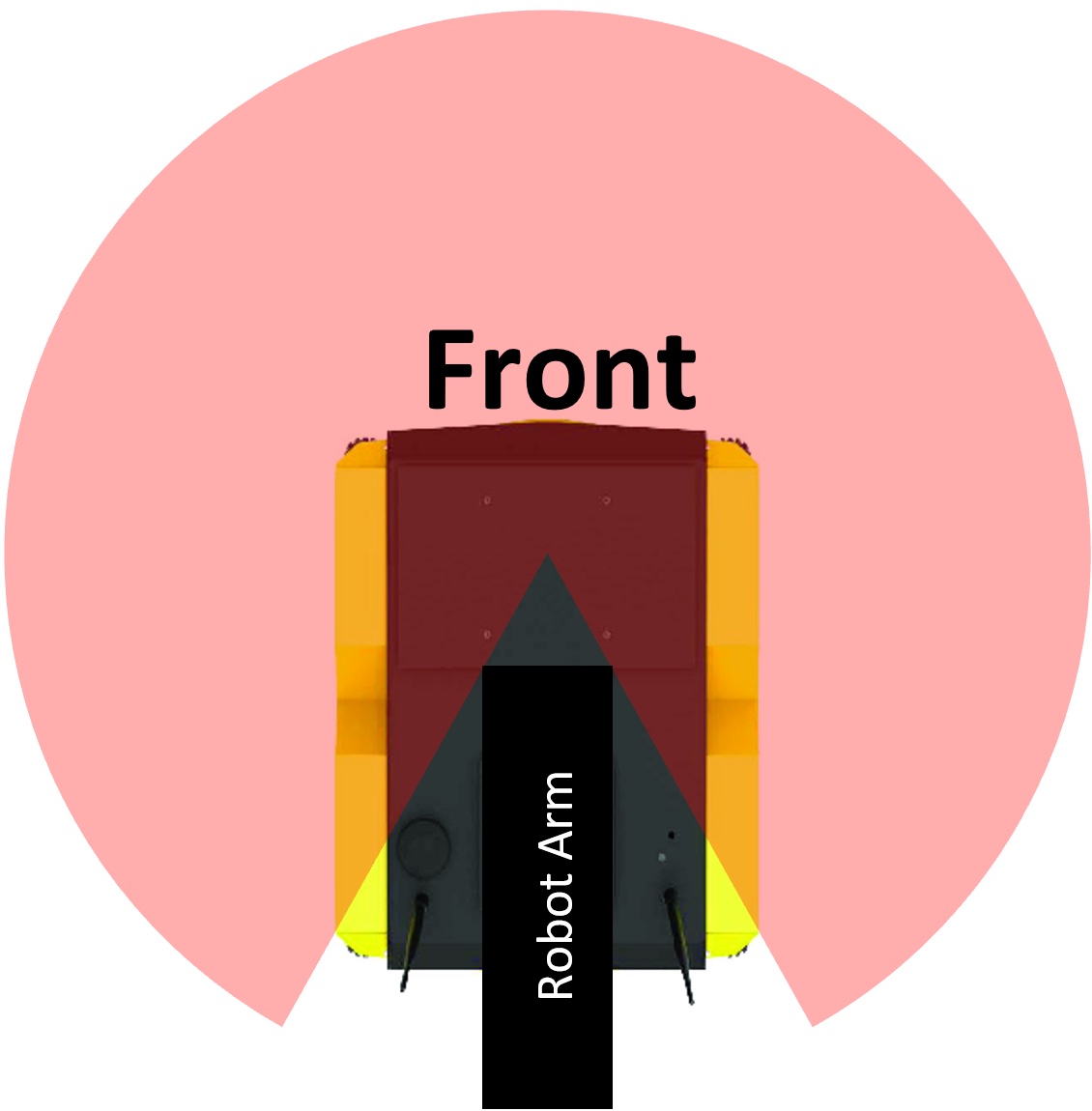}
     \caption{Top-down view of the Jackal Robot. The laser scan is occluded due to the robot arm yielding a field of view of 300° as depicted with the red circle.}
     \label{fig:gg_jackal}
\end{figure}

\textbf{Gazebo Gym} is similar to the Point Robot Navigation meaning that it shares the same task and constraint set $\mathcal{C}$ but resembles a realistic environment. The agent is a Jackal differential drive robot. Furthermore, the environment size is increased from 1x1 to 8x8 m. Similarly, the goal region is enlarged to 1 m in the easy and  to 0.3 m in the hard setting of Gazebo Gym. The observation of the agent is given as
\begin{equation}
    \label{eq:gg_observation}
    \vec{x}_{observation} = [\vec{d}, \sin(\psi), \cos(\psi), v_\text{long}, \dot{\psi}, \vec{d}_\text{laser} ],
\end{equation}
where $\vec{d}$ denotes the vector from the current robot position to the goal, $\psi$ the yawing angle of the robot and $v_\text{long}$ the longitudinal velocity, assuming zero slip. The environment is encoded in $\vec{d}_\text{laser}$ which is a 120-dimensional vector containing the 1D range measurements of each laser ray with small additive Gaussian noise. The laser scan has a field of view of 300° as depicted in Fig.~\ref{fig:gg_jackal}. We first have been experiencing with the sparse reward setting in Eq.~\ref{eq:boc_reward} in Point Robot Navigation. However, both SAC and PPO baseline performed poorly with success rates around 10\% at 250k steps. To account for the more challenging dynamics of a differential drive robot, we implemented the dense reward
\begin{equation}
    \label{eq:gg_reward}
    \begin{aligned}
    r_\textrm{constraint\_free} = 
    & \begin{cases}
    40 & \text{if } ||\vec{d}||_2 < 0.1 \\
    -0.1  ||\vec{d}||_2 & \text{else} 
    \end{cases} \,, \\
    r_\textrm{constraint} = & -20 \,.
    \end{aligned}
\end{equation}

The term $-0.1 ||\vec{d}||_2$ is a dense signal that directly connects the observation of the robot in Eq.~\ref{eq:gg_observation} to the reward. 
The environment is implemented in Gazebo~\cite{gazebo}. For PPO, we train 9 agents simultaneously, whereas for SAC, only one agent is used. The sampling time is 0.1s. At each iteration, the agent starts at a random position, and every forty iterations, a new set of obstacles is randomly spawned. In the hard mode, the agent is subject to Gaussian additive action noise. An example of the environment is shown in Fig.~\ref{fig:gg}.

\subsection{Additional Experimental Results}

\begin{figure*}
     \centering
     \includegraphics[width=0.9\textwidth]{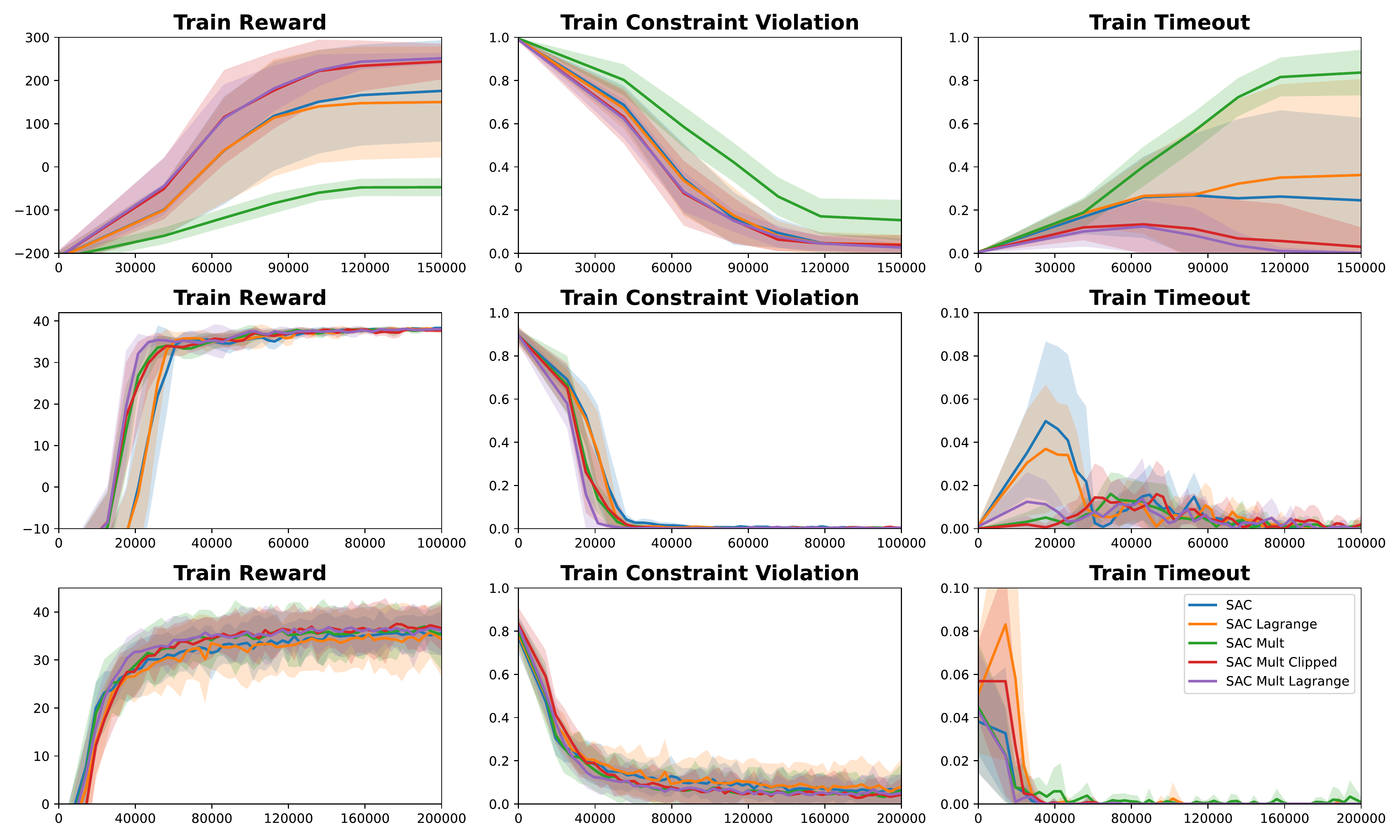}
     \caption{SAC training curves. Top to bottom: Lunar Lander Safe, Point Robot Navigation, Gazebo Gym.}
     \label{fig:train_sac}
 \end{figure*}

\begin{figure*}
     \centering
     \includegraphics[width=0.9\textwidth]{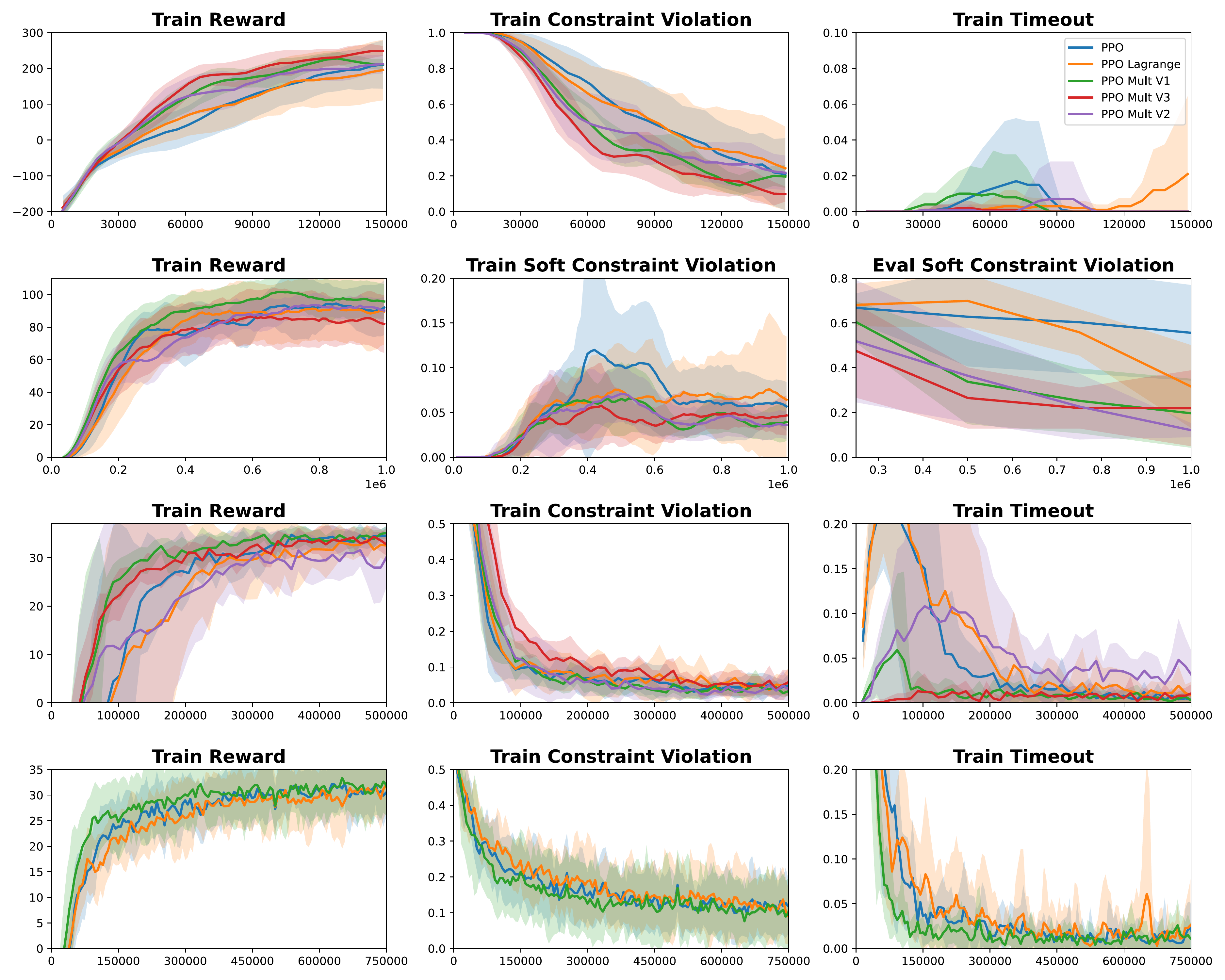}
     \caption{PPO training curves. Top to bottom: Lunar Lander Safe, Car Racing Safe, Point Robot Navigation, Gazebo Gym.}
     \label{fig:train_ppo}
 \end{figure*}

We provide the training curves of SAC in Fig.~\ref{fig:train_sac} and of PPO in Fig.~\ref{fig:train_ppo}. Compared to the evaluation results in Table~1 and 2 of the main paper, the algorithms violate the constraints more frequently in training. However, this is expected since the randomness of the training policy can cause the agent to reach potentially dangerous states. Furthermore, in Fig.~\ref{fig:train_ppo}, we provide the soft constraint violation rates in Car Racing Safe. Interestingly, the agent drives faster with the deterministic policy such that the soft constraint is violated more frequently in evaluation.

\begin{figure*}
    \centering
    \includegraphics[width=0.75\textwidth]{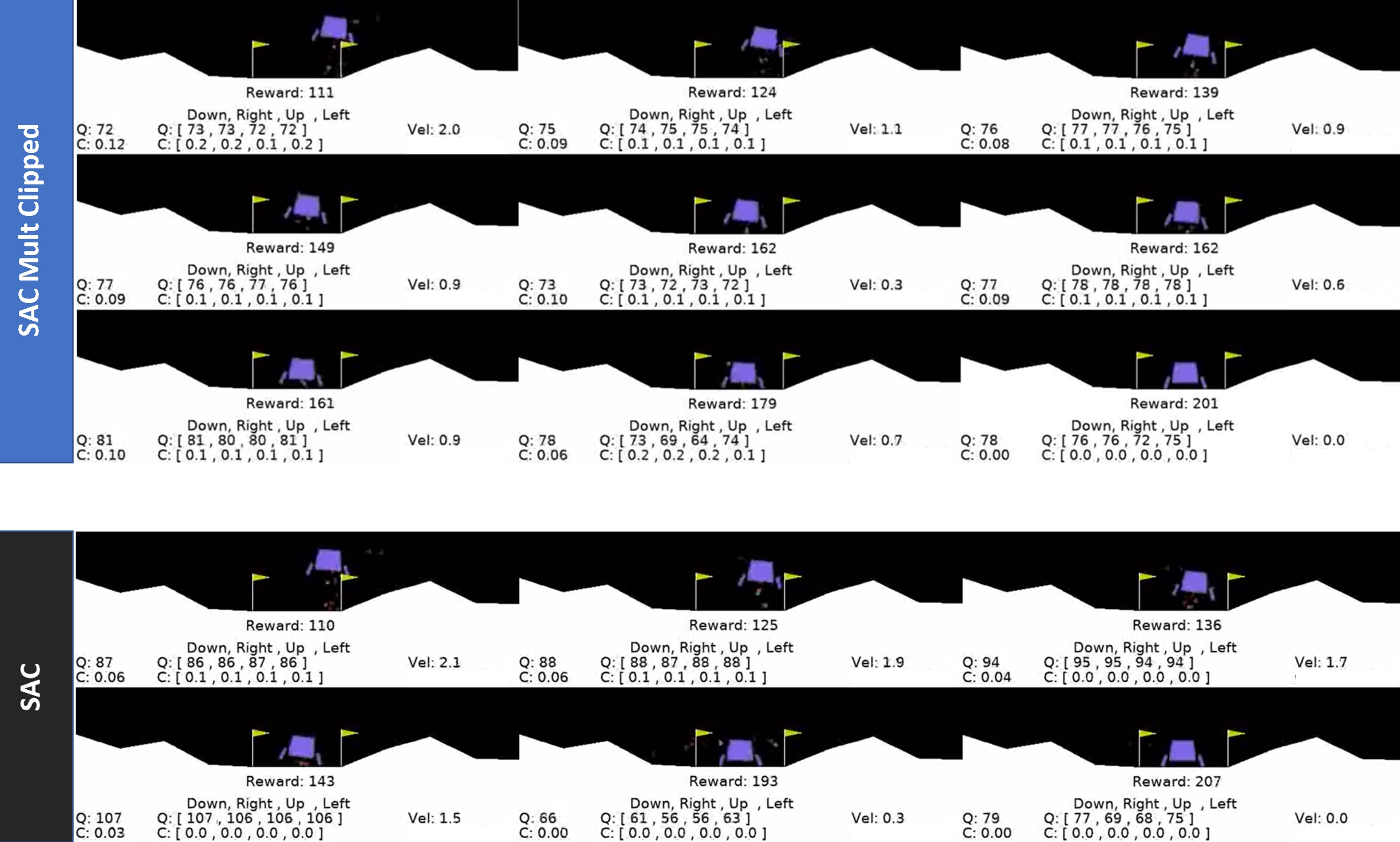}
    \vspace{-1ex}
    \caption{Qualitative comparison of SAC vs SAC Mult Clipped on seed six after 300k steps on Lunar Lander Safe. We depict every fifth frame of a one-episode video evaluation. The images have to be read from left to right, top to bottom.  In each frame, the first column with Q and C denotes the estimated value and constraint violation probability of the next suggested action.  The second column shows the action values and the constraint violation probabilities for basic actions like going up, left, right or down.  The last column depicts the current downwards velocity of the lander.}
    \vspace{-3ex}
    \label{fig:ll_qualitative_sac}
\end{figure*}

We furthermore present qualitative results for Lunar Lander Safe, Car Racing Safe and Gazebo Gym. In Fig.~\ref{fig:ll_qualitative_sac}, we compare the landing of SAC Mult Clipped against SAC baseline on Lunar Lander Safe. In the first row, the flights of baseline and Mult Clipped agent look similar. However, in the third frame, we see that the downward velocity of the baseline lander is 1.7 m/s, which is much higher than 0.9 m/s of the multiplicative version. When continuing with the baseline plot, second row, first image, we see that the lander slows down by 0.2 m/s and uses its right leg to establish ground contact. This makes the agent decelerate from 1.5 to 0.3 m/s, shown in the next frame. In contrast, the multiplicative agent lands more conservatively. Starting from the second row, the next five frames show the agent decelerating. Shortly before landing, the agent has a velocity of 0.7 m/s and lands with both legs simultaneously. This is much safer from a real-world perspective: Slowing down from 0.7→0 m/s with two legs compared to 1.5→0.3 m/s with one leg like the baseline.

\begin{figure*}
    \centering
    \includegraphics[width=0.75\textwidth]{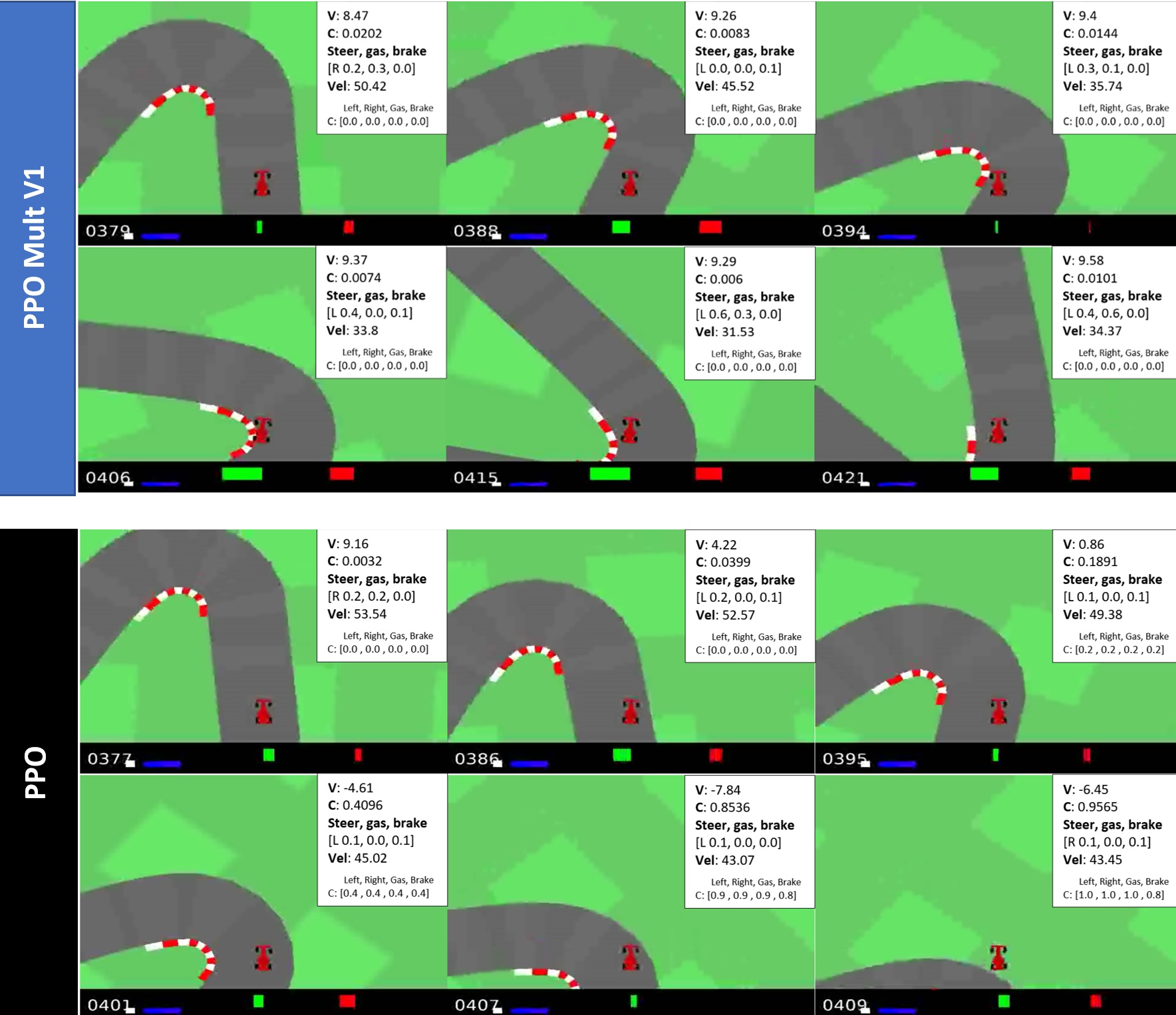}
    \vspace{-1ex}
    \caption{Qualitative comparison of PPO base vs PPO Mult V1 on seed six after 1M steps. We depict every fifth frame of a one-episode video evaluation. The images have to be read from left to right, top to bottom. In each frame, the first two rows with V and C denote the estimated value and constraint violation probability at the current state with the next suggested action. The third and fourth row depict the next suggested action. The last row contains the constraint violation probabilities for basic actions like steering left, right, accelerating and braking.}
    \label{fig:cr_video}
\end{figure*}

Next, we regard the first frame of Car Racing Safe in Fig.~\ref{fig:cr_video}. Both the multiplicative and baseline agents steer to the right to open up the corner. When proceeding to the third frame, we can see that the baseline agent has opened up the corner more but also has a higher velocity of 49 km/h than the multiplicative agent. On the other side the multiplicative agent slows down to 36 km/h and cuts the inside. Continuing with frame four, we see that the velocity of the baseline agent is still high at 45 km/h. This limits the agent's ability to steer to the left without losing traction. At this point, the safety critic estimates a crashing probability of 40\% (note that we train a safety critic also in the baseline version for visualization purposes, but to not use it in any way for the policy training and stop all possible gradients). As we proceed, the agent cannot make the corner and leaves the track. We now regard the multiplicative policy. In the fourth frame, we can see that the velocity is low at 34 km/h. This allows the agent to steer more aggressively to the left. In the next frame, the agent suggests further steering to the left but also starts to accelerate. This is a real-world racing technique where one starts to accelerate after the apex of a corner to increase traction.

\begin{figure*}
    \centerline{\includegraphics[width=0.98\textwidth]{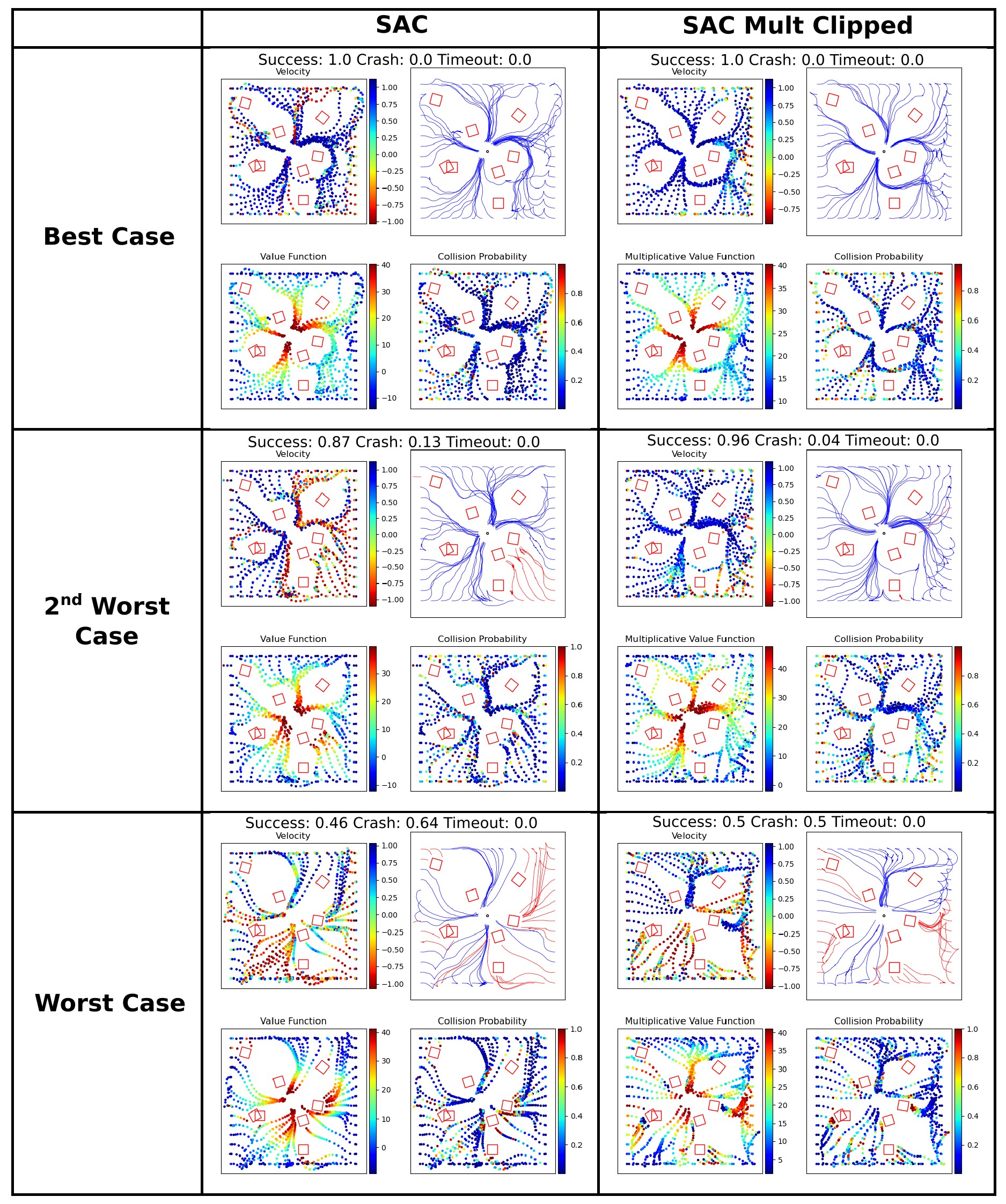}}
    \caption{Qualitative results on Gazebo Gym with SAC after 200k steps. The Velocity, Value Function, and Collision Probability plots show the corresponding metric at each third trajectory point. The best and worst case distinction are with respect to the best and worst success rates out of ten seeds.}
    \label{fig:gg_img_sac}
\end{figure*}

Finally, we switch to the Gazebo Gym results shown in Fig.~\ref{fig:gg_img_sac} and start with the best case, where both SAC baseline and Mult Clipped achieve a success rate of 100\%. Furthermore, the trajectories of the multiplicative version seem smoother and more natural compared to the baseline. Another difference is that our approach strictly drives with the front forward, whereas the baseline sometimes drives back first. This is shown in the velocity plots, where a velocity smaller than zero means driving back-first. This is potentially dangerous since the agent has a 60° blindspot in the back as shown in Fig.~\ref{fig:gg_jackal}. Note that the behavior of driving back or front first is an emerging behavior that was not incentivized by the reward.

In the second-worst case, the behavior of occasionally driving back first causes the baseline to crash, as can be seen in the bottom right corner of the trajectory plot. On the other hand, our approach drives head-first and achieves a success rate of 96\%. For the 4\% of trajectories in which the agent crashes, there is an anomaly in the reward critic: the estimated return is overly optimistic, especially close to obstacles. The worst case is obtained with seed two, where the combination of driving back first and anomalies in the value functions cause baseline and multiplicative agents to crash every second trajectory. However, since the second-worse case achieves significantly higher performance, we consider this second seed an outlier. 

\subsection{Additional Ablation Studies}

\begin{figure*}
     \centering
     \begin{subfigure}[b]{\textwidth}
         \centering
         \includegraphics[width=\textwidth]{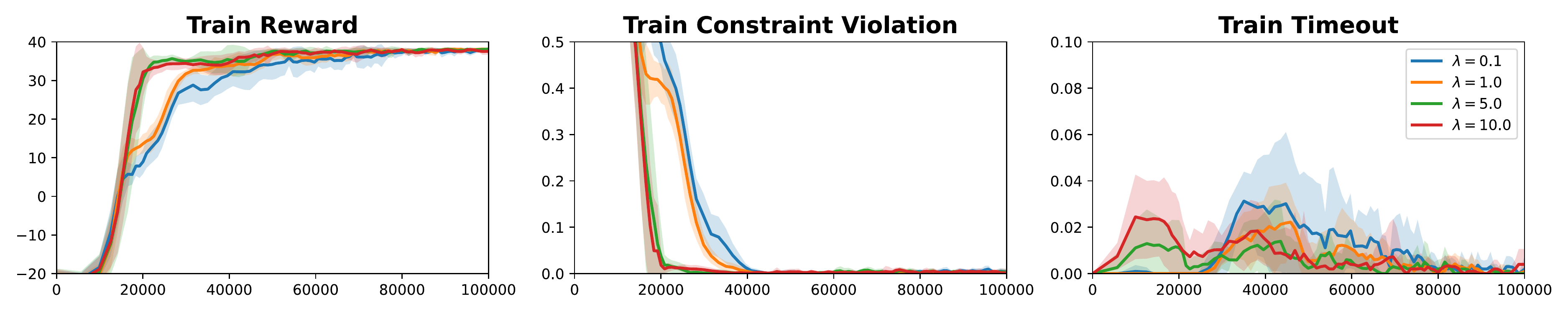}
         \caption{Varying $\lambda_\text{init}$ for SAC Mult Lagrange, $\gamma_c = 0.8$}
         \label{fig:sac_lagrange}
     \end{subfigure}
     \hfill
     \begin{subfigure}[b]{\textwidth}
         \centering
         \includegraphics[width=\textwidth]{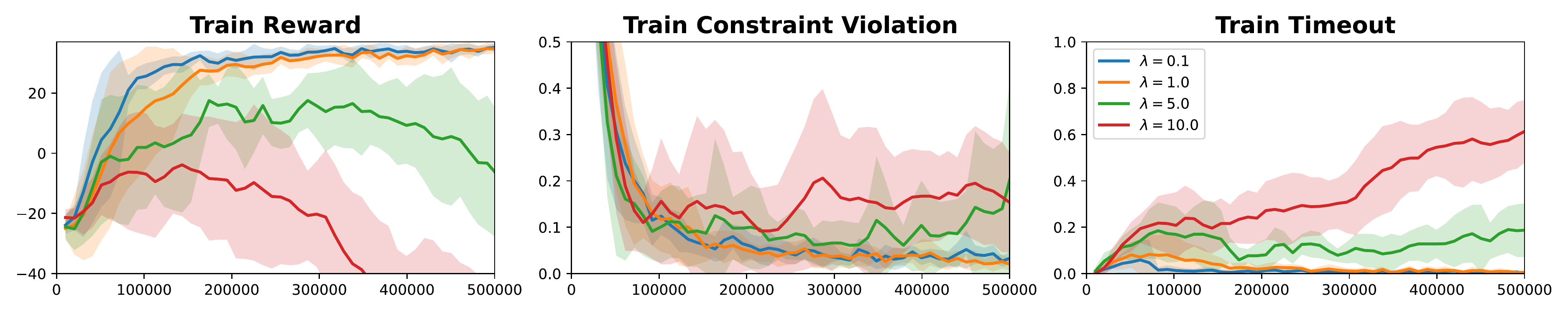}
         \caption{Varying $\lambda_\text{init}$ for PPO Mult V1, $\gamma_c = 0.8$.}
         \label{fig:ppo_lagrange}
     \end{subfigure}
     \hfill
     \begin{subfigure}[b]{\textwidth}
         \centering
         \includegraphics[width=\textwidth]{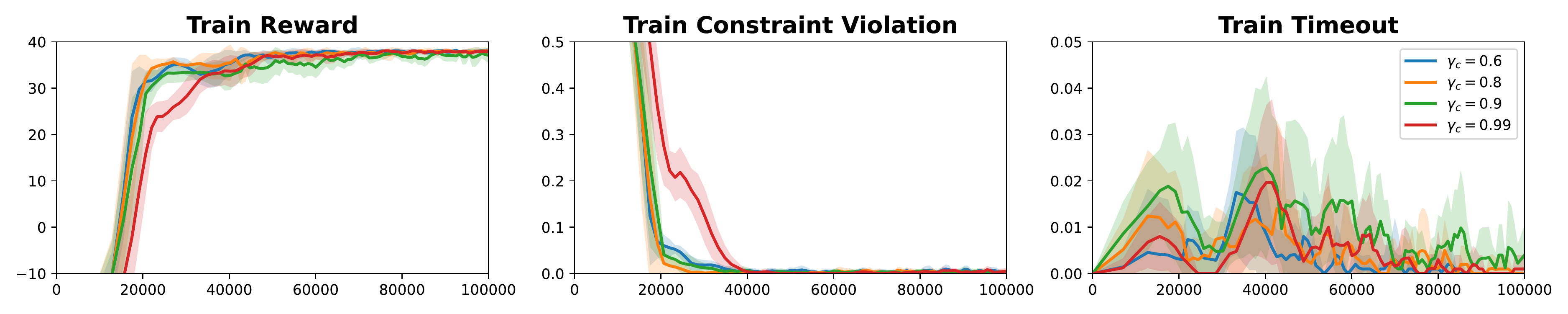}
         \caption{Varying $\gamma_c$ for SAC Mult Lagrange, $\lambda_\text{init} = 5.0$.}
         \label{fig:sac_gamma}
     \end{subfigure}
     \hfill
     \begin{subfigure}[b]{\textwidth}
         \centering
         \includegraphics[width=\textwidth]{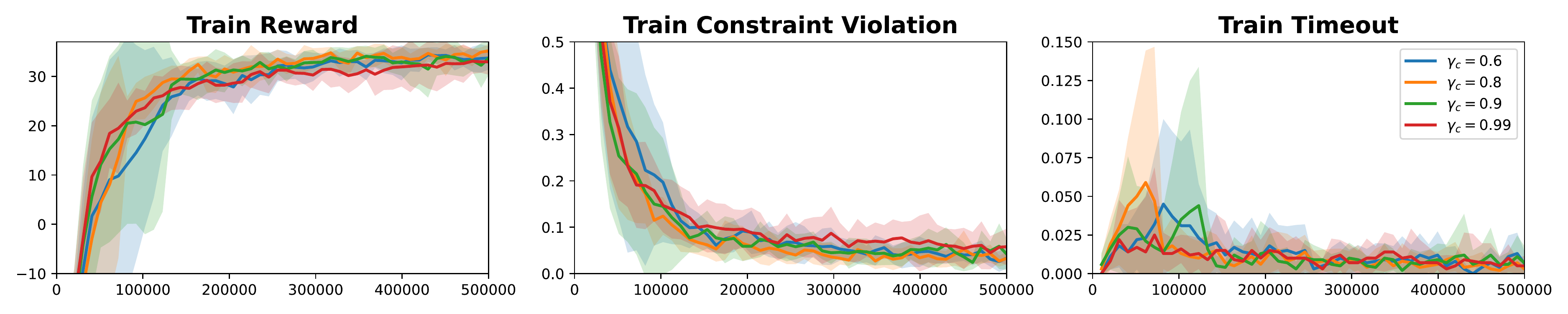}
         \caption{Varying $\gamma_c$ for PPO Mult V1, $\lambda_\text{init} = 0.1$.}
         \label{fig:ppo_gamma}
     \end{subfigure}
        \caption{Ablation Experiments in Point Robot Navigation}
        \label{fig:ablation}
\end{figure*}

Finally, we want to show the effects of the choice of initial Lagrange multiplier $\lambda_\text{init}$ and safety discount factor $\gamma_c$ which is depicted in Fig.~\ref{fig:ablation}. For SAC in Fig.~\ref{fig:sac_lagrange}, the different choices of the initial Lagrange multiplier do not significantly affect the performance at convergence but can result in different sample efficiencies. However, Fig.~\ref{fig:ppo_lagrange} shows PPO is more sensitive where multipliers with magnitude five and higher cause training instabilities with rising timeout rates. Note that in the stable-baselines3 implementation~\cite{stable-baselines3} upon which we build our code, the advantage in PPO is normalized whereas the reward in SAC is not. This means that the different magnitudes of the Lagrange multiplier have a greater effect on PPO than on SAC in Point Robot Navigation. 

Furthermore, we experiment with different safety discount factors $\gamma_c$ shown in Fig.~\ref{fig:sac_gamma} and \ref{fig:ppo_gamma}. Both for SAC and PPO the safety discount factor seems to have little effect on the performance. This can be explained by the fact that the ``dynamics" of the point robot allow for an instantaneous change of direction such that a short safety horizon with low $\gamma_c$ is sufficient for obstacle avoidance.

\clearpage

\begin{algorithm}[H]
    \caption{SAC Multiplicative}
    \label{alg:safe_sac}
\begin{algorithmic}[1]
    \State \textbf{Init}: policy parameters $\theta$, $\bar{Q}$-function parameters $\xi_1$, $\xi_2$, \textcolor{blue}{Safety critic parameters $\psi_1$, $\psi_2$}, empty replay buffer $\mathcal{D}$.
    \State Set target parameters equal to main parameters $\xi_{\text{targ},i} \leftarrow \xi_i,$ \textcolor{blue}{$\psi_{\text{targ},i} \leftarrow \psi_i$}, for $i\in\{1,2\}$
    
    \Repeat
        \State Observe state $s$ and sample action $a \sim \pi_{\theta}(\cdot|s)$.
        \State Observe next state $s'$ and done signal $d$.
        \State \textcolor{blue}{Observe clipped reward $\bar{r}$ and constraint cost $r_c$.}
        \State Store $(s,a,\bar{r},$ \textcolor{blue}{$r_c$}$,s',d)$ in replay buffer $\mathcal{D}$.
        \State Reset environment states \textbf{if} $s'$ is terminal.
        \If{it's time to update}
            \For{$j$ in range(however many updates)}
                \State \multiline{Randomly sample a batch of transitions from $\mathcal{D}$, $B = \{ (s,a,\bar{r},$ \textcolor{blue}{$r_c$}$,s',d) \}$ .}
                \State \multiline{Compute targets $y (\bar{r},s',d)$ for $\bar{Q}$-functions:
                \begin{equation*}
                    \bar{r} + \gamma (1-d) (\min_{i} \bar{Q}_{\xi_{\text{targ},i}} (s', \tilde{a}') - \alpha \log \pi_{\theta}(\tilde{a}'|s'))
                \end{equation*}
                where $\tilde{a}' \sim \pi_{\theta}(\cdot|s')$
                }
                \State \multiline{\textcolor{blue}{Compute targets $y_c (r_c,s',d)$ for safety critic:}
                \textcolor{blue}{
                \begin{equation*}
                    r_c + \gamma_c (1-d) (\max_{i} \Psi_{\psi_{\text{targ}, i}} (s', \tilde{a}') ), 
                \end{equation*} 
                where $\tilde{a}' \sim \pi_{\theta}(\cdot|s')$
                }
                }
                \State \multiline{ Update $\bar{Q}$-functions for $i\in\{1,2\}$:
                \begin{equation*}
                    \nabla_{\xi_i} \frac{1}{|B|}\sum_{b\in B} \left( \bar{Q}_{\xi_i}(s,a) - y(\bar{r},s',d) \right)^2,
                \end{equation*}
                where $b = (s,a,\bar{r},r_c,s',d)$.
                }
                \State \multiline{ \textcolor{blue}{Update safety critic for $i\in\{1,2\}$:
                \begin{equation*}
                    \nabla_{\psi_i} \frac{1}{|B|}\sum_{b \in B} \text{BCE} \left( \Psi_{\psi_i}(s,a) , y_c(r_c,s',d) \right),
                \end{equation*}
                where BCE denotes the binary cross-entropy.}
                }            
                \State \multiline{Update policy by one step of gradient ascent:
                \begin{equation*}
                    \nabla_{\theta} \frac{1}{|B|}\sum_{s \in B} \textcolor{blue}{\text{$Q_{\text{mult},\xi,\psi}(s, \tilde{a}_{\theta})$}} - \alpha \log \pi_{\theta} ( \tilde{a}_{\theta} | s),
                \end{equation*}
                where $\tilde{a}_{\theta}(s)$ is sampled from $\pi_{\theta}(\cdot|s)$ via reparametrization trick.
                }
                \State \multiline{Update target networks:
                \begin{align*}
                    \xi_{\text{targ},i} \leftarrow \rho \cdot \xi_{\text{targ}, i} + (1-\rho) \xi_i \, , \\
                   \text{ \textcolor{blue}{$ \psi_{\text{targ},i} \leftarrow \rho_c \cdot \psi_{\text{targ}, i} + (1-\rho_c) \psi_i \, ,$}}
                \end{align*}
                where $i\in\{1,2\}$.
                }
            \EndFor
        \EndIf
    \Until{convergence}
\end{algorithmic}
\end{algorithm}

\begin{algorithm}[H]
    \caption{PPO Multiplicative}
    \label{alg:ppo}
\begin{algorithmic}[1]
    \State \textbf{init}: policy parameters $\theta_0$, value function parameters $\xi_0$, \textcolor{blue}{safety critic parameters $\psi_{1,0}$, $\psi_{2,0}$, maximum unsafety probability target $c_{\max}$, and Lagrange multiplier $\lambda$.} 
    \For{$k = 0,1,2,...$}
    \State \multiline{Collect set of trajectories ${\mathcal D}_k = \{\tau_i\}$ by running policy $\pi_k = \pi(\theta_k)$ in the environment.}
    \State \multiline{Compute \textcolor{blue}{clipped} rewards-to-go $\hat{R}_t$ and \textcolor{blue}{constraint cost-to-go $\hat{C}_t$} .}
    \State \multiline{Compute advantage estimates, \textcolor{blue}{$\hat{A}_\text{mult}$}  with current value function $\bar{V}_{\xi_k}$ and \textcolor{blue}{safety critic $\Psi_{ \psi_k}$}.}
    \State \multiline{\textcolor{blue}{Approximate $\hat{\Phi}^{\pi_\theta} (s):=\mathbb{E}_{a_\theta \sim \pi_\theta}[\Psi(s, a_\theta) - c_{\max}]$ the expectation in the PPO Mult objective by sampling from the policy
    \begin{equation*}
    \hat{\Phi}^{\pi_\theta} (s) \approx \frac{1}{N} \sum_{i=0}^N \max_{j=1,2} \Psi_{\psi_j}(s, a_{\theta_i}) - c_{\max},
    \end{equation*}
    where $a_{\theta_i} \sim \pi_\theta$. 
    }}
    \State \multiline{Update the policy $\theta$ by maximizing the PPO-Clip Mult objective:
        \begin{align*}
         \frac{1}{|{\mathcal D}_k| T} \sum_{\tau \in {\mathcal D}_k} \sum_{t=0}^T \min\Big(
            \frac{\pi_{\theta}(a_t|x_t)}{\pi_{\theta_k}(a_t|x_t)}  \text{\textcolor{blue} { $A_\text{mult}^{\pi_{\theta_k}}(x,a)$}},  \\
            g(\epsilon,  \text{\textcolor{blue} { $A_\text{mult}^{\pi_{\theta_k}}(x,a)$}}) 
        \Big)  - \text{\textcolor{blue}{$ \lambda \hat{\Phi}^{\pi_\theta}(s_t)$}},
        \end{align*}
        typically via stochastic gradient ascent with Adam. }
    \State \multiline{ \textcolor{blue}{Update the Lagrange Multiplier by}
    \begin{equation*}
    \text{\textcolor{blue}{$
    \lambda \leftarrow \max\left (0,   \lambda + \alpha \frac{1}{|{\mathcal D}_k| T} \sum_{\tau \in {\mathcal D}_k} \sum_{t=0}^T \hat{P}_c^{\pi_\theta}(x_t)
    \right )$}} ,
    \end{equation*}
    \textcolor{blue}{where $\alpha$ is a learning rate.}
   }
    \State \multiline{Fit value function using mean-squared error:
    \begin{equation*}
    \xi_{k+1} = \arg \min_{\xi} \frac{1}{|{\mathcal D}_k| T} \sum_{\tau \in {\mathcal D}_k} \sum_{t=0}^T\left( \bar{V}_{\xi} (s_t) - \hat{R}_t \right)^2
    \end{equation*}
    }
    \State \multiline{\textcolor{blue}{Update safety critic parameter $\psi_{i, k+1}$ by minimizing the binary cross entropy:
    \begin{align*}
       \frac{1}{|{\mathcal D}_k| T} \sum_{\tau \in {\mathcal D}_k} \sum_{t=0}^T \text{BCE} \left( \Psi_{\psi_i}(x,a) , \hat{C}_t \right),
    \end{align*}
    where $i \in \{1,2\}$.
     }}
    \EndFor
\end{algorithmic}
\end{algorithm}


\end{document}